\documentclass{article}

\usepackage{arxiv}

\usepackage[utf8]{inputenc} % allow utf-8 input
\usepackage[T1]{fontenc}    % use 8-bit T1 fonts
\usepackage{hyperref}       % hyperlinks
\usepackage{url}            % simple URL typesetting
\usepackage{booktabs}       % professional-quality tables
\usepackage{amsfonts}       % blackboard math symbols
\usepackage{nicefrac}       % compact symbols for 1/2, etc.
\usepackage{microtype}      % microtypography
\usepackage{lipsum}		% Can be removed after putting your text content
\usepackage{graphicx}
\usepackage{natbib}
\usepackage{doi}

\title{Test-time Adversarial Defense with\\ Opposite Adversarial Path and High Attack Time Cost}

\author{Cheng-Han Yeh \& Kuanchun Yu \& Chun-Shien Lu \\
Institute of Information Science \\
Academia Sinica \\
Taipei, Taiwan (ROC) \\
\texttt{\{jerryyeh,mikeyu1012,lcs\}@iis.sinica.edu.tw} \\
}

\renewcommand{\headeright}{}
\renewcommand{\shorttitle}{Test-time Adversarial Defense with Opposite Adversarial Path and High Attack Time Cost}

\hypersetup{
pdftitle={A template for the arxiv style},
pdfsubject={q-bio.NC, q-bio.QM},
pdfauthor={David S.~Hippocampus, Elias D.~Striatum},
pdfkeywords={First keyword, Second keyword, More},
}

\usepackage{booktabs}       % professional-quality tables
\usepackage{amsfonts}       % blackboard math symbols
\usepackage{nicefrac}       % compact symbols for 1/2, etc.
\usepackage{microtype}      % microtypography
\usepackage{xcolor}         % colors
\usepackage{adjustbox}
\usepackage{multirow}
\usepackage{caption}
\usepackage{amssymb}
\usepackage{rotating}

\usepackage{tablefootnote}
\usepackage{multirow}
\usepackage{arydshln} %负责画虚线的包
\usepackage{graphicx}
\usepackage{amsmath}
\usepackage[normalem]{ulem}
\usepackage{enumitem}
\usepackage[rightcaption]{sidecap}

\usepackage{pifont}

\usepackage{algorithm}
\usepackage{algorithmic}
\usepackage{lipsum}
\usepackage{filecontents}
\usepackage{xcolor}

\makeatletter
\renewcommand\fs@ruled{\def\@fs@cfont{\bfseries}\let\@fs@capt\floatc@ruled
  \def\@fs@pre{{\color{black}\hrule height.8pt depth0pt \kern2pt}}%
  \def\@fs@post{{\color{black}\kern2pt\hrule\relax}}%
  \def\@fs@mid{{\kern2pt\color{black}\hrule\kern2pt}}%
  \let\@fs@iftopcapt\iftrue}
\makeatother

    \makeatother

\newcommand{\cL}{{\mathcal L}}

\newcommand{\cN}{{\mathcal N}}
\newcommand{\cO}{{\mathcal O}}

\newcommand{\cS}{{\mathcal S}}

\newcommand{\EE}{\mathbb{E}}
\newcommand{\NN}{\mathbb{N}}

\renewcommand{\a}{\alpha}  
 
\newcommand{\g}{\gamma}
 
\newcommand{\e}{\epsilon}

\newcommand{\m}{\mu}

\newcommand{\s}{\sigma}

\newcommand{\p}{\phi}

\newcommand{\SG}{\Sigma}
\newcommand{\vp}{\varphi} 
\newcommand{\ve}{\varepsilon}

\newcommand*{\sgn}{\operatorname{sign}}

\newcommand{\argmin}{\displaystyle\operatorname*{argmin}} %
\newcommand\norm[1]{\left\Vert #1\right\Vert} %norm [#] %
\newcommand\abs[1]{\left\vert #1\right\vert} %absolute value [#] %

\renewcommand*{\int}{\varint}
\def\code#1{\texttt{#1}}

\def\AA{AutoAttack}
\def\AP{adversarial perturbation}

\begin{document}
\maketitle

\begin{abstract}
    Deep learning models are known to be vulnerable to adversarial attacks by injecting sophisticated designed perturbations to input data. Training-time defenses still exhibit a significant performance gap between natural accuracy and robust accuracy. In this paper, we investigate a new test-time adversarial defense method via diffusion-based recovery along opposite adversarial paths (OAPs). We present a purifier that can be plugged into a pre-trained model to resist adversarial attacks. Different from prior arts, the key idea is excessive denoising or purification by integrating the opposite adversarial direction with reverse diffusion to push the input image further toward the opposite adversarial direction. For the first time, we also exemplify the pitfall of conducting AutoAttack (Rand) for diffusion-based defense methods. Through the lens of time complexity, we examine the trade-off between the effectiveness of adaptive attack and its computation complexity against our defense. Experimental evaluation along with time cost analysis verifies the effectiveness of the proposed method.
\end{abstract}

% keywords can be removed

\section{Introduction}
\label{sec:intro}
\subsection{Background}
It has been well known that deep learning models are vulnerable to adversarial attacks by injecting (imperceptible) adversarial perturbations into the data that will be input to a neural network (NN) model to change its normal predictions \cite{Athalye2018,Carlini2019,Soups-CVPR2023, Frosio2023,Goodfellow2015,gowal2021improving,Madry2018,ASD-WACV2023}.
Please also see \cite{PY-AAAI2023} for a recent review on the adversarial robustness of deep learning models.
It can be found from the literature that adversarial attacks defeat their defense counterparts easily and rapidly, and there is still a gap between natural accuracy and robust accuracy. 

The study of adversarial defense in resisting adversarial attacks can be divided into two categories: (1) Adversarial training/Training-time defense \cite{gowal2021improving,GAT-CVPR2023,AAD-CVPR2023,Suzuki2023,wang2019improving,wang2023better,wu2020adversarial,zhang2019theoretically}; and (2) Input pre-processing/Test-time defense \cite{alfarra2022combating,Chen-transduction2022,hill2020stochastic,ho2022disco,nie2022diffusion,wang2022guided,wu2022guided,yoon2021adversarial}.
Adversarial training utilizes adversarial examples derived from the training data to enhance the robustness of the classifier. 
Despite the effort in training-time defense, we do see (\href{https://robustbench.github.io/}{RobustBench} \cite{Robustbench2021}) there is still a remarkable gap between natural accuracy and robust accuracy.

Different from the training-time defense paradigm, in this paper, we propose a new test-time adversarial defensive method by pre-processing data in a way different from the prior works.
It is a kind of purifier and serves as a plug-and-play module that can be used to improve the robustness of a defense method once our module is incorporated as a pre-processor. 
Specifically, the formulation of processing the input data is derived as: $\min_{\p,\,\theta}\EE\left[\max_{x'\in B(x)}\cL((f_\p\circ g_\theta)(x'),y)\right]$,
where $x'$ denotes the adversarial example corresponding to clean image $x$ with label $y$, $B(\cdot)$ is the threat model, $f_\p$ is the image classifier parameterized by $\p$, and $g_\theta$ is a pre-processor.

A key to test-time defense is the design of pre-processor or denoiser ({\em e.g.}, $g_\theta$), which aims at denoising an adversarial example to remove the added perturbations. 
Intuitively, the goal is to have the denoised image as close to the original one 
so as to achieve perceptual similarity. 

\subsection{Related Works}
\label{Sec: Related Work}
We introduce representative test-time adversarial defense methods \cite{alfarra2022combating,ho2022disco,hill2020stochastic,yoon2021adversarial,nie2022diffusion,wang2022guided,wu2022guided} that share the same theme as our method.
Please also see Sec. \ref{Related Works: Supp} in the Supplementary for details of \cite{hill2020stochastic,yoon2021adversarial,wang2022guided,wu2022guided}. 

In \cite{alfarra2022combating}, a defense method is proposed by connecting an anti-adversary layer with a pre-trained classifier $f_\p$. Given an input image $x$, it will be first sent to the anti-adversary layer for generating anti-adversarial perturbation $\g$ by solving an optimization problem. 
As the name implies, in most cases, the direction $\g$ will be opposite to the direction of adversarial perturbation. The resultant purified image $x+\g$ is then used for classification. 

DISCO \cite{ho2022disco} is proposed as a purification method to remove adversarial perturbations by localized manifold projections. The author implemented it with an encoder and a local implicit module, which is leveraged by the architecture called LIIF \cite{chen2021learning,chen2019learning}, where the former produces per-pixel features and the latter uses the features in the neighborhood of query pixel for predicting the clean RGB value.

In DiffPure \cite{nie2022diffusion}, given an input (clean or adversarially noisy), the goal is to obtain a relatively cleaner version through a series of forward and reverse diffusion processes. Moreover, a theoretical guarantee is derived that, under an amount of Gaussian noise added in the forward step, adversarial perturbation may be removed effectively. 
This is independent of the types of adversarial perturbations, making DiffPure defend against unseen attacks.

Recently, the robustness of diffusion-based purifiers is considered overestimated. \cite{lee2023robust} provides recommendations for robust evaluation, called {\em surrogate process}, and shows that defense methods may be defeated under the surrogate process. \cite{kang2024diffattack} proposes DiffAttack, a new attack against diffusion-based adversarial purification defenses, that can overcome the challenges of attacking diffusion models, including vanishing/exploding gradients, high memory costs, and large randomness. The use of a segment-wise algorithm allows attacking with much longer diffusion lengths than previous methods.

Although the aforementioned purification-based adversarial defense methods show promising performance in resisting adversarial attacks, \cite{croce2022evaluating} argues that their evaluations are ineffective in two aspects: (i) Incorrect use of attacks or (ii) Attacks used for evaluation are not strong enough.
However, the authors also mentioned 
test-time defense complicates robustness evaluation because of its complexity and computational cost, which impose even more computations for the attackers.

\subsection{Motivation}\label{subsection-motivation}
Let us take image classification as an example, where clean/natural accuracy is the classification accuracy for benign images and robust accuracy is measured for adversarial samples.
However, we argue that ``perceptually similar’’ does not mean adversarial robustness as it is not guaranteed to entirely remove the {\AP}s such that the residual perturbations still have an impact on changing the prediction of a learning model.
On the contrary, we propose to purify the input data along the direction of opposite adversarial paths (OAPs) excessively, as shown in Fig. \ref{new ref data}.

\begin{table}[tb]
\centering
% \vspace{-10pt}
\begin{minipage}{.4\linewidth}
  \centering
    \centering
    \includegraphics[width=4.1cm]{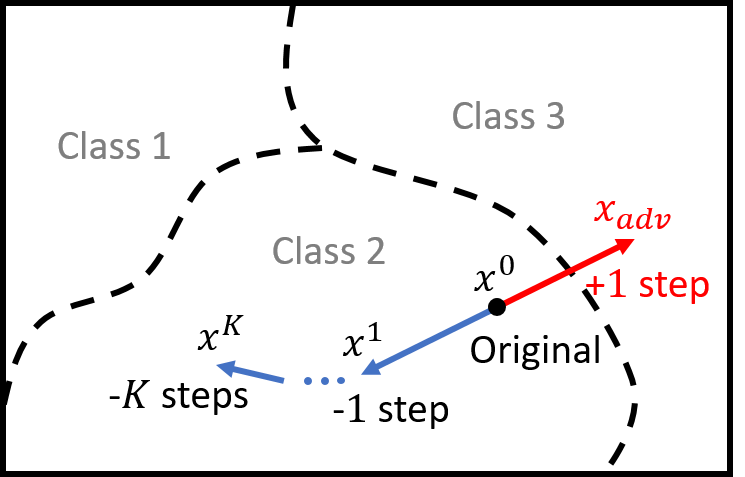}
    \vspace{10pt}
    \captionof{figure}{\small Concept diagram of new reference point generation via $K$ consecutive purifications along opposite adversarial paths (OAPs).}
    \label{new ref data}
\end{minipage}%
\hspace{1.5pt}
\begin{minipage}{.59\linewidth}
\centering
  \centering
  \normalsize
  \renewcommand\arraystretch{1.3}
  \resizebox{0.66\textwidth}{!}{
  \begin{tabular}{@{}ccc@{}}
  % \begin{tabular}{p{15cm}p{15cm}p{15cm}}
    \toprule
    -$K$ Steps & Clean Accuracy (\%) & Robust Accuracy (\%) \\
    \midrule
    0 & 95.16 & 0.18 \\
    -1 & 89.94 & 6.18 \\
    -3 & 98.56 & 83.64 \\
   -5 & 99.63 & 96.78 \\
%    7 & 99.9 & 99.19 \\
    -10 & 99.96 & 99.86 \\
%    15 & 100 & 99.98 \\
    -20 & 100.00 & 100.00 \\
    \bottomrule
  \end{tabular}}
    \vspace{10pt}
  \captionof{table}{\small Pre-processing the training dataset by adding $K$ steps of $-adv$ (via PGD (\cite{madry2017towards}); see Fig. \ref{new ref data}) and feeding to a non-defense classifier (ResNet-18 (\cite{he2016deep})) pre-trained on CIFAR-10 (\cite{krizhevsky2009learning}) for testing. %Clean Acc and Robust Acc denote classification accuracy for benign and adversarial samples, respectively.
  %Note that there is no pre-processing when 
  $K=0$ indicates original data.}%Overall, this is merely a toy example of demonstrating our intuition as manipulated training dataset was used for testing.}
  \label{Tab: -ADV(Input)}
    
\end{minipage}
% \vspace{-20pt}
\end{table}

% \begin{table}[t]
% \centering
% \begin{minipage}{1.\linewidth}
%   \centering
%     \centering
%     \includegraphics[width=4cm]{figs/new_ref_data.png}
%     \captionof{figure}{\small Conception of new reference point generation via $K$ consecutive purifications along opposite adversarial paths (OAPs).}
%     \label{new ref data}
% \end{minipage}%
% \hspace{1.5pt}
% \begin{minipage}{1.\linewidth}
% \centering
%   \centering
%   \normalsize
%   \renewcommand\arraystretch{1.3}
%   \resizebox{0.4\textwidth}{!}{
%   \begin{tabular}{@{}ccc@{}}
%     \toprule
%     -$K$ Steps & Clean Accuracy (\%) & Robust Accuracy (\%) \\
%     \midrule
%     0 & 95.16 & 0.18 \\
%     -3 & 98.56 & 83.64 \\
%    -5 & 99.63 & 96.78 \\
%     -10 & 99.96 & 99.86 \\
%     -20 & 100.00 & 100.00 \\
%     \bottomrule
%   \end{tabular}}
%   \captionof{table}{\small Pre-processing the training dataset by adding $K$ steps of $-adv$ (via PGD \cite{madry2017towards}; see Fig. \ref{new ref data}) and feeding to a non-defense classifier (ResNet-18 \cite{he2016deep}) pre-trained on CIFAR-10 \cite{krizhevsky2009learning} for testing. 
%   $K=0$ indicates original data.}
%   \label{Tab: -ADV(Input)}
    
% \end{minipage}
% \vspace{-20pt}
% \end{table}

Conceptually, if we add the {\AP} along the opposite direction of Projected Gradient Descent (PGD) \cite{madry2017towards}, denoted as ``$-adv$,'' to a given data, robust accuracy can be improved.
To gain an insight that excessive denoising (more than one step along the opposite gradient) is advantageous in resisting attacks, 
a simple experiment was conducted by moving each data point $x$ to the new position $x^K$ through $K$ iterations of opposite adversarial perturbation, according to the ground truth label and classifier.
Given each kind of $x^K$, the accuracy change is illustrated in Table \ref{Tab: -ADV(Input)}.

Moreover, 
motivated by \cite{croce2022evaluating}, our defense method also aims to complicate the computation of adaptive adversarial attacks.

\subsection{Contributions} 
Different from prior works, the concept of OAP can be incorporated into any training scheme of purifiers, and the OAP-based purifier can also become a part of modules in other defense processes. For instance, OAP-based purifiers can provide additional directions within reverse diffusion, whereas diffusion models alone \cite{song2020score} (baseline model in DiffPure) only provide direction to generate images. Unlike the traditional purification methods, we do not use \textit{classifier-generated labels} (\textit{e.g.} Anti-Adv \cite{alfarra2022combating}) in our baseline purifier during testing. 
On the contrary, combining the proposed baseline purifier with the reverse diffusion process provides reference directions pointing to a safer area during the purification process. 

Contributions of this work are summarized as follows:

(1) We are first to present the idea of excessive denoising along the opposite adversarial path (OAP) as the baseline purifier for adversarial robustness (Sec. \ref{opposite adv}). 
%  \item 
(2) We integrate the OAP baseline purifier and conditional reverse diffusion as a sophisticated adversarial defense that can be interpreted as moving purified data toward the combination of directions from the score-based diffusion model and baseline purifier (Sec. \ref{Conditional RDP}).
%  \item 
(3) To complicate the entire defense mechanism by complicating the computation overhead of adaptive attacks accordingly, we study a double diffusion path cleaning-based purifier (Sec. \ref{DPC}). This creates a trade-off between the attack effectiveness and attack computation. % budget.
%  \item 
(4) For the first time, we exemplify the pitfall of conducting AutoAttack (Rand) for diffusion-based adversarial defense methods (Sec. \ref{Sec: GGA}).

\section{Preliminary}
%In this section, 
%We provide some notations frequently used throughout this paper. To make this paper self-contained, adversarial attacks and diffusion models are briefly introduced.
%we introduce the threat models, common settings used in evaluation, and related defense components.

\subsection{Basic Notation} 
\label{notation}

In the paper, $x$ denotes an input image, $\hat x$ denotes a recovered image or overly denoised/purified image, $x_{adv}$ denotes an adversarial image, $y$ is a ground-truth label of $x$, $\hat{y}$ is a prediction, $g_\theta$ is a purifier, and $f_\p$ is a pre-trained classifier.

For the diffusion model, the forward process is denoted by $q(\cdot|\cdot)$ and the backward/reverse process is denoted by $p_{\theta}(\cdot|\cdot)$ with parameter $\theta$. 
For $t\in[0,T]$, $x_t$ represents an image at time step $t$ during the forward / reverse diffusion process. Usually, $x_0$ is a clean image and $x_T\sim\cN(0,I)$. 
%For the conditional generation, we denoted the reference image in the condition as $c$.

%\subsection{Adversarial Attack}
For the adversarial attack, it modifies the input image $x$ by adding to it adversarial perturbation $\delta$ by calculating the gradient of loss according to information leakage of pre-trained NN $f_\p$ without changing $\p$, causing $f_\p$ to classify incorrectly. 
According to the leakage level, there are roughly two types of attacks. Please see Sec. \ref{Sec: AA} in the Supplementary.
%for details.

\subsection{Diffusion Models}
Since the diffusion model \cite{sohl2015deep,ho2020denoising,song2019generative,song2020score} is a baseline model in diffusion-based purifiers, to make this paper self-contained, please refer to Sec. \ref{Sec: DMs: Supp} of Supplementary for a brief introduction to the diffusion model.

%Taking logarithm in (\ref{equ condition}) gives the compact form:
%\begin{align}
%    \log(p_\vp(x_{t-1}|x_t)p_\p(y|x_{t-1}))\approx\log(z)+C,
%\end{align}
%where $z\sim\cN(\m_\vp(x_t)+\SG_\vp(x_t) g,\SG_\vp(x_t))$ and $g=\nabla_{x_t}\log p_\p(y|x_t)\mid_{x_t=\m_\vp}$.

%\begin{align*}
%    z&\sim\cN(\m_\vp(x_t)+\SG_\vp(x_t) g,\SG_\vp(x_t)),\\
%    g&=\nabla_{x_t}\log p_\p(y|x_t)\mid_{x_t=\m_\vp}.
%\end{align*}
\section{Proposed Method}
\label{method}
\begin{figure}[ht]
        % \vspace{-10pt}
        \centering
        \includegraphics[width=0.9\linewidth]{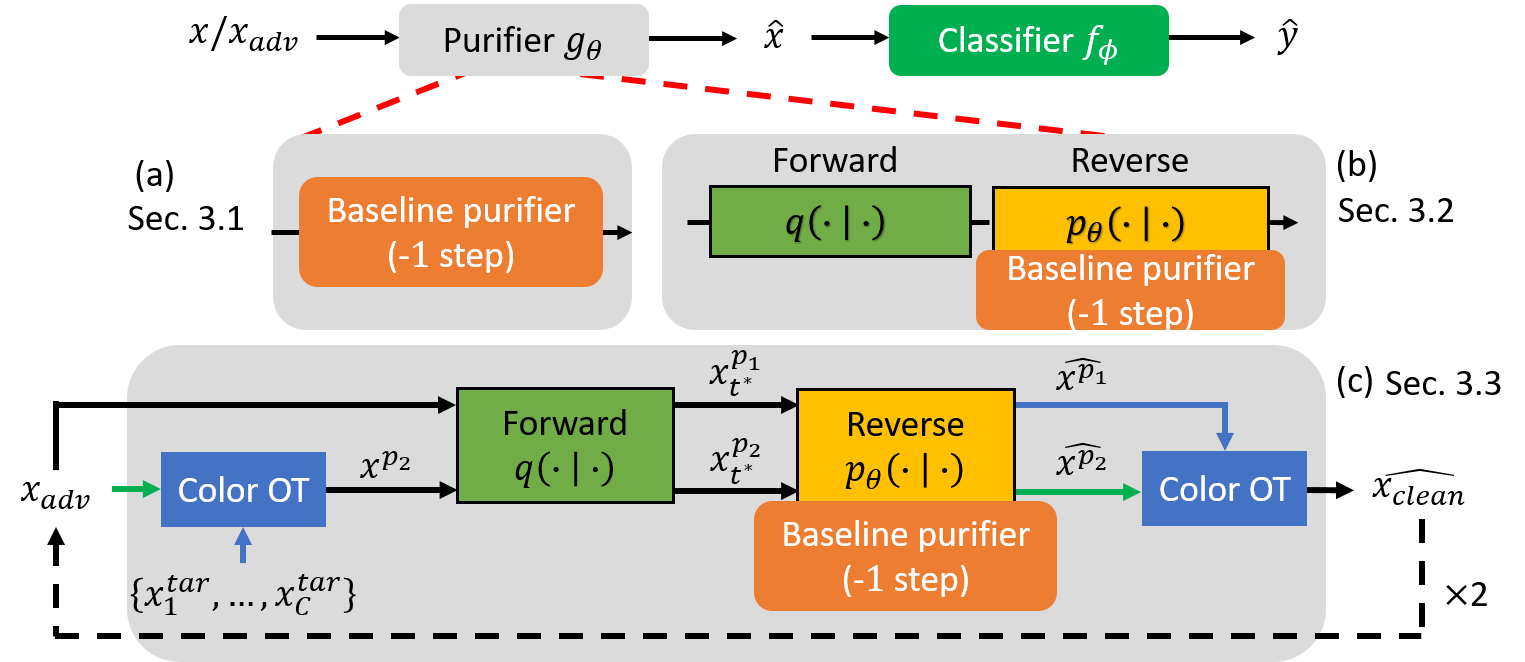}
        \caption{\small Flowchart of our method. The purifier (gray block) can be one of (a)-(c), where (a) is the proposed baseline purifier, (b) shows the combination of baseline purifier and reverse diffusion, and (c) expands (b) with two diffusion paths. In (c), $x_1^{tar}, \ldots, x_C^{tar}$ are obtained via Eq.\! (\ref{eq: data generation rule}) from fixed $C$ images with one image per class. The image in front of Color OT with green/blue arrow is called the source/target image. $x^{p_2}$ is defined in Eq.\! (\ref{eq: color ot}).}
        \label{defense flow chart}
\end{figure}

We describe the proposed test-time adversarial defense method with its flowchart illustrated in Fig. \ref{defense flow chart}. 

\subsection{Baseline Purifier: Opposite Adversarial Path (OAP)} \label{opposite adv}
Given a classifier model $f_\p$ parameterized by $\p$, a loss function $\cL(x, y, \p)$, and a pair of data $(x, y)$, the adversarial attack can be computed as:
%\begin{align}
    $x_{adv}=\Pi_{x+\cS}(x+\a\sgn(\nabla_x\cL(x,y,\p)))$,
%\end{align}
where $\cS$ is the set that allows the perceptual similarity between natural and adversarial images.
This iterative process aims to find the adversarial image $x_{adv}$ that maximizes the loss function. 

On the other hand, the opposite direction of each iteration points to minimize the loss. Assume now we get an ordinary noisy input $x_{adv}=x+\delta$ with $\norm{\delta}_p\le\e_p$ via image processing, a denoiser can push the denoised input close to $x$ within a non-perceptual distortion.
%, where classifier can readily predict.
Nevertheless, if the noisy input $x_{adv}$ is a sophisticated design via adversarial attack, it is too early to claim, depending on the perceptual similarity between $x$ and $x_{adv}$, that the denoised image can be free from being affected by adversarial perturbations. 
We argue that if we properly push the denoised image further away from the decision boundary, the downstream classifier can still successfully classify the input since the direction we push points to a lower loss area on the input-loss surface, as illustrated in Fig. \ref{new ref data}.
Also note that the {\em plug-and-play} module lies under the setting that the baseline purifier is only trained on a given attack ({\em e.g.}, PGD-$\ell_\infty$-7), which is independent of the attacks ({\em e.g.}, PGD-$\ell_\infty$-40, AutoAttack, and BPDA+EOT) used in testing. 
%We follow this scheme to train and evaluate our baseline purifier. 
In addition, the diffusion model is pre-trained (Sec. \ref{Conditional RDP}) and does not involve adversarial examples during its training.

\subsubsection{New Reference Point Generation} 
Previous test-time defense methods with a plug-and-play fashion take $x_{adv}$ as an input and generate the predicted ``clean'' image $\hat x$. In our scenario, we want to move a few steps further.
Starting from the clean image $x$, ground-truth label $y$, parameter $\p$ and loss function $\cL$ of classifier $f$,
we can generate a new reference point $x^K$ \textit{for training} by:
%the following formula:
\begin{align}\label{eq: data generation rule}
    x^k=\Pi_{x^{k-1}+\cS}(x^{k-1} - \a\sgn(\nabla_{x^{k-1}}\cL(x^{k-1},y,\p))),
\end{align}
for $1\leq k\leq K$, where $x^0=x$. If we iterate Eq.\! (\ref{eq: data generation rule}), we can get a series of data, $x^1, x^2, \ldots, x^K$, as illustrated in Fig. \ref{new ref data}.
%and Fig. \ref{data prep and training}.}
%and bottom-left of Fig. \ref{data prep and training}.}

\subsubsection{Baseline Purifier Training}
In traditional denoising, the goal is to train a purifier that produces a denoised output $\hat{x}$ from the adversarial input $x_{adv}$, denoted as $x_{adv} \mapsto \hat{x}$, such that $\hat{x}$ and $x$ can be as similar as possible in terms of, say, $\ell_p$-norm. 
We, instead, train the purifier to produce $\widehat{x^K}$ from $x_{adv}$ that further points toward the opposite adversarial attack direction. 
We call the resultant $\widehat{x^K}$ an excessively-denoised image and the model $g_\theta$ that moves data along the opposite adversarial path (OAP) the ``baseline purifier.''

In practice, we train a baseline purifier %{\color{blue}$g_\th(\cdot)$} 
using data pairs $\{(x_{adv}, x^K)\}$ with a certain number of opposite steps $K \in \NN$, where $x^K$ is generated by Eq.\! (\ref{eq: data generation rule}). 
The training procedure of $g_\theta$ is to minimize:
\begin{align}\label{eq: baseline puri training}
    \theta^\ast=\argmin_\theta \norm{g_\theta(x_{adv})-x^K}_1,
\end{align}
where $g_\theta$ can be any existing defense methods ({\em e.g.}, DISCO \cite{ho2022disco}).
The results of training on different opposite steps 
%{\color{blue}$K$} 
are shown in Table \ref{tab: disco vs 3Attacks} with respect to PGD-$\ell_\infty$ \cite{madry2017towards}, AutoAttack (AA) \cite{AutoAttack2020}, and BPDA \cite{Athalye2018}.
We can observe that the idea of the new reference point indeed improves DISCO. 
Specifically, when $K=1$, the robust accuracy can be improved greatly, but it decreases as $K$ goes larger. 
The results are somewhat inconsistent with those in Table \ref{Tab: -ADV(Input)}. 
The reason we conjecture is that the experiment presented in Table \ref{Tab: -ADV(Input)} was conducted using the ground-truth label to move data step-by-step, but it is not in Table \ref{tab: disco vs 3Attacks}. 
Besides, as $K$ increases, the distance that needs to push the data increases as well, similar to the effect of large step size in gradient descent, \textit{e.g.}, coarse gradient estimation near the decision boundary.
Therefore, based on the empirical observations, we will empirically set $K=1$ for learning the opposite direction of an adversarial attack during training.

On the other hand, we will later demonstrate that OAP is a powerful module readily to be incorporated with existing adversarial defenses ({\em {e.g.}}, DISCO) in improving both the clean and robust accuracy.
%in Tables \ref{Table: Robustness Evaluation Non-adaptive} and  \ref{Table: Robustness Evaluation Adaptive}.
% \begin{table}[]
% \begin{tabular}{c|ll}
% \hline
% \multicolumn{1}{|l|}{\multirow{2}{*}{\textit{-K} steps}} & \multicolumn{2}{l|}{Non-adaptive PGD-$\ell_\infty$}                         \\ \cline{2-3} 
% \multicolumn{1}{|l|}{}                  & \multicolumn{1}{l|}{Clean Acc (\%)} & \multicolumn{1}{l|}{Robust Acc (\%)} \\ \hline
%                                         &                       &                       \\
%                                         &                       &                       \\
%                                         &                       &                       \\
%                                         &                       &                      
% \end{tabular}
% \end{table}

% % Please add the following required packages to your document preamble:
% % \usepackage{multirow}
% \begin{table}[]
% \begin{tabular}{c|cc}
% \hline
% \multirow{2}{*}{\textit{-K} steps} & \multicolumn{2}{c}{Non-adaptive PGD-$\ell_\infty$} \\ \cline{2-3} 
%                    &   Clean Acc (\%)        &     Robust Acc (\%)     \\ \hline
%             0       &           &          \\
%             -1      &           &          \\
%             -3      &           &          \\
%             -7      &           &          \\ \hline
% \end{tabular}
% \end{table}

\begin{table*} 
\centering
\normalsize
\renewcommand\arraystretch{1.3}
%\resizebox{0.3\textwidth}{!}{
  \resizebox{0.8\textwidth}{!}{\begin{tabular}{@{}c|cc@{}}
    \hline
    \multirow{2}{*}{\textit{-K} steps} & \multicolumn{2}{c}{Non-adapt PGD-$\ell_\infty$ / ResNet-18} \\ 
    \cline{2-3} 
     & Clean Acc (\%) & Robust Acc (\%) \\
    \hline
    0 & 89.57 & 73.13 \\
    -1 & 90.71 & 86.10 \\
    -3 & 89.53 & 82.02 \\
    -7 & 89.33 & 56.21 \\
    \hline
  \end{tabular}
  \hspace{0.2in}
  % \hfill
  \begin{tabular}{@{}cc@{}}
    \hline
    \multicolumn{2}{c}{Non-adapt AA / WRN-28-10} \\ \hline
    Clean Acc (\%) & Robust Acc (\%) \\
    \hline
    89.00 & 85.00 \\
    91.66 & 88.79 \\
    90.23 & 86.14 \\
    89.58 & 69.04 \\
    \hline
  \end{tabular}
  \hspace{0.2in}
  % \hfill
  \begin{tabular}{@{}cc@{}}
    \hline
    \multicolumn{2}{c}{BPDA / VGG16} \\ \hline
    Clean Acc (\%) & Robust Acc (\%) \\
    \hline
    88.38 & 47.37 \\
    89.26 & 56.34 \\
    87.78 & 59.94 \\
    88.42 & 52.31 \\
    \hline
  \end{tabular}}
%  }
  \caption{\small Evaluation of DISCO trained with the relation between new reference point %(in different steps) 
  and {\AP}s by PGD attack generated in ResNet-18. Entire CIFAR-10 testing dataset was used. (Left) Attack: Non-adaptive PGD-$\ell_\infty$. Test model: ResNet-18. (Middle) Attack: Non-adaptive AutoAttack (AA). Test model: WRN-28-10 \cite{WRN}. (Right) Attack: BPDA. Test model: VGG16 \cite{VGG16}.}
  \label{tab: disco vs 3Attacks}
\end{table*}

%the mapping cannot capture the {\color{red}nature} of  (\ref{eq: data generation rule}) since there is a difference between step-by-step update, ${\color{blue}x_{adv}\to} x^0\to x^1\to\cdots\to x^K$, and mapping, ${\color{blue}x_{adv}}\to x^K$.
%While the model is in testing, we purify the input data in an iterative manner, as indicated in (\ref{eq: data generation rule}).

\subsection{Diffusion-based Purifier with OAP Prior}\label{Conditional RDP}
%Combine directions and connection to SGLD
In Sec. \ref{opposite adv}, we have witnessed the merit of baseline purifier based on OAP %(new reference point generation plugged into a defense method ({\em e.g.}, DISCO \cite{ho2022disco})) 
in improving robustness against adversarial attacks. 
%\textcolor{blue}{This is said to be realized by a kind of purifier design ({\em e.g.}, denoising vs. excessive denoising) or OAG prior.
%The attackers, however, may access the available information from the framework of purifier+classifier to design adaptive attacks as a stronger countermeasure against this defense mechanism.
This data moving trick also motivates us to study how to incorporate OAP prior and diffusion models as a stronger adversarial defense.
%plays an important role in pushing input data to different locations. 
%To further understand how it affects the purification process, we compare the diffusion-based purifier, which learns the distribution of training dataset, with different reverse processes we propose, including conditional reverse and mixture of purification results.}
%with baseline purifier described in Sec. \ref{opposite adv}.
%The attackers, however, may access the available information from the framework of purifier+classifier to design adaptive attacks as a stronger countermeasure against this defense mechanism.

%\textcolor{blue}{For the conditional reverse process with diffusion-based purifier}, 
We first propose to integrate the idea of opposite adversarial paths with the reverse diffusion process ({\em e.g.}, guided diffusion \cite{dhariwal2021diffusion}, ILVR \cite{choi2021ilvr}, and DDA \cite{gao2023back}) to achieve a similar goal of pushing the input image further toward the opposite adversarial direction. 
More importantly, for each step in the reverse diffusion process, the purifier is used to provide a direction that points to $x^K$.
%, \sout{and this also increases the computation complexity for the attackers to generate adaptive attack examples.}

To this end, according to Eq.\! \eqref{equ condition} of guided diffusion described in Sec. \ref{Sec: DMs: Supp} in Supplementary, 
% To this end, we first recall guided diffusion \cite{dhariwal2021diffusion}, that given a label $y$ as the condition and (\ref{Eq: Reverse-Diffusion}), the conditional reverse process is specified as:
% \begin{align}\label{eq: guided reverse}
%     p_\th(x_0,\ldots,x_{T-1}|x_T,y)
%     =\prod_{t=1}^Tp_\th(x_{t-1}|x_t,y).
% \end{align}
% % For simplicity(???why), we may consider one term 
% To solve the reversed output in (\ref{eq: guided reverse}), $\th$ is decomposed into two terms as $\th=\vp\cup\p$ to form separate models:
% \begin{align} \label{equ condition}
%     p_\th(x_{t-1}|x_t,y)
%     =Zp_\vp(x_{t-1}|x_t)p_\p(y|x_{t-1}),
% \end{align}
% where $Z$ is a normalization constant. 
by taking logarithm and gradient with respect to $x_{t-1}$ \cite{dhariwal2021diffusion}, we can derive
\begin{align} \label{condition grad}
    &\nabla_{x_{t-1}} \log p_\theta(x_{t-1}|x_t,y) \nonumber\\
    &= \nabla_{x_{t-1}} \log p_\vp(x_{t-1}|x_t) + \nabla_{x_{t-1}} \log p_\p(y|x_{t-1}),
\end{align}
where $t$ denotes the diffusion time step.
%, as described in Sec. \ref{notation}.
Based on Langevin dynamics, we get a sampling chain on $x_{t-1}$ as:
\begin{align} \label{LD update}
    x_{t-1} \leftarrow x_t + \nabla_{x_{t-1}} \log p_\vp(x_{t-1}|x_t),
\end{align}
where we get the first direction (specified by $p_\vp$) of moving to $x_{t-1}$.
However, if we want to generate $x_{t-1}$ by moving along the direction given $x_t$ and $y$, we have to introduce the second direction (specified by $p_\p$) to move to $x_{t-1}$ given condition $y$ based on Eq.\! \eqref{condition grad}. Hence, we add $\nabla_{x_{t-1}} \log p_\p(y|x_{t-1})$ in the sampling chain \eqref{LD update} as:
\begin{align}
    x_{t-1} \leftarrow x_t 
    &+ \nabla_{x_{t-1}} \log p_\vp(x_{t-1}|x_t) \nonumber\\
    &+ \nabla_{x_{t-1}} \log p_\p(y|x_{t-1}),
\label{Eq: Second Direction}
\end{align}
where the last two terms are the same as the RHS of Eq.\! \eqref{condition grad}.
%This results in an update rule similar to Stochastic gradient Langevin dynamics (SGLD) \cite{welling2011bayesian}.
Note that the second term can be approximated by a model $\mathbf{\e_{\vp}}(\cdot)$ that predicts the noise added to the input.
According to (11) in \cite{dhariwal2021diffusion}, it can be used to derive a score function as:
\begin{align}
    \nabla_{x_{t-1}} \log p_\vp(x_{t-1}|x_t) = -\frac{\mathbf{\e_{\vp}}(x_{t-1})}{\sqrt {1-\bar{\alpha}_t}},
\end{align}
where $\bar{\a}_t=\prod_{s=1}^t(1-\beta_s)$.

Different from previous works, if $y$ in the third term of Eq.\! (\ref{Eq: Second Direction}) is replaced with the new reference point $x^K$, as described in Eq.\! (\ref{eq: data generation rule}) of Sec. \ref{opposite adv}, then the term becomes $\nabla_{x_{t-1}} \log p_\p(x^K|x_{t-1})$ and represents how to move along the direction to $x^K$ given $x_{t-1}$. This can be set by 
\begin{align}
    \widehat{x^K} \leftarrow g_\theta(x_{t-1});\nonumber\\ %\nonumber\\
    \nabla_{x_{t-1}} \log p_\p(x^K|x_{t-1}) \approx \eta(\widehat{x^K} - x_{t-1}),
    \label{Eq: Double Directions}
\end{align}
where $\eta$ is the step size and $g_\theta(\cdot)$ is the purifier (see Sec. \ref{opposite adv}) that can approximate the mapping of $x_{adv}\to x^K$.
Hence, the purification process can be interpreted as moving toward the combination of directions from the score-based diffusion model \cite{nie2022diffusion,song2019generative,song2020score} and baseline purifier $g_\theta(\cdot)$. 

%\textcolor{blue}{Another option is to mixup different purifiers in that we simply average the outputs from the diffusion-based purifier and the baseline purifier $g_\th(\cdot)$. The performance will be shown later in Table \ref{Table: Robustness Evaluation}, indicated in ``+ OAG''.}

\subsubsection{Connecting the OAP Prior with Diffusion}

We are aware that the base purifier has to operate in the domain the same as that in the diffusion reverse process, {\em i.e.}, they deal with different inputs with noises at different scales. 
However, according to Eq.\! (\ref{eq: baseline puri training}), the baseline purifier only takes inputs that are adversarially perturbed. 
Hence, during the training of baseline purifier, we randomly add different scales of noise to the input data so that the base purifier can accommodate the different noise scales in the reverse diffusion process, denoted as:
\begin{align}\label{eq: connect oag prior and diffusion}
    \theta^\ast_n=\argmin_\theta \mathbb{E}_{p_{data}(x_{adv})}\mathbb{E}_{p_{\sigma_t}(\tilde{x}|x_{adv})}\norm{g_\theta(\tilde{x})-x^K}_1,
    \end{align}
where $t$ is uniformly chosen from $0\ldots t^\ast$, $\s_t$ is the corresponding noise scale at diffusion time step $t$, and $\tilde{x}$ is the perturbed data according to the diffusion process. We replace the baseline purifier $g_\theta$ in Eq.\! (\ref{Eq: Double Directions}) in Sec. \ref{Conditional RDP} with this purifier $g_{\theta_n}$.

\subsection{Diffusion Path Cleaning-based Purifier}\label{DPC}
%In addition to the baseline purifier in Sec. \ref{opposite adv} and conditional reverse diffusion-based purifier in Sec. \ref{Conditional RDP}, 
%\sout{As described before, we can view the baseline purifier (Sec. \ref{opposite adv}) and the conditional reverse diffusion-based purifier (Sec. \ref{Conditional RDP}) as the constraints that force data to move to a safer zone.
%({\em e.g.}, $x^K$).
%However, under the adaptive adversarial setting, the adversary can easily obtain an attack direction to attack through the purifier and even the reverse diffusion process.}
In this section, we describe how to further utilize other gradients from different constraints to modify/move our samples toward specific directions.
Moreover, the goal is to complicate the entire framework of purifier+classifier so as to complicate the computation of adaptive attacks accordingly while maintaining  comparable clean and robust accuracy. 
%Specifically, as shown in ILVR \cite{choi2021ilvr} and DDA \cite{gao2023back}, a low-pass filter or a distance measure, which minimizes the distance between an image in the reverse diffusion process and a low-passed image from the reference image, will be exploited to guide image generation.
%as in ILVR\cite{choi2021ilvr}. 
%, just like the modification of the input image in a -1 adversarial direction.}
We first conduct a test to verify whether such a framework could be affected by such an attack.
%on CIFAR-10.

%\begin{table}[tb]
%\centering
\begin{figure}[t]
% \vspace{-10pt}
\centering
  \begin{minipage}[t]{.45\textwidth}
    \centering
      \centering      
        \includegraphics[width=0.8\linewidth]{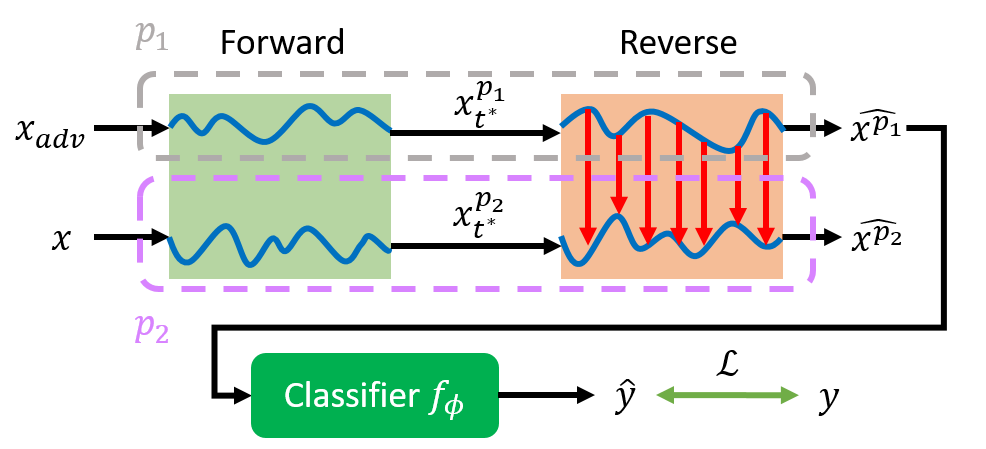}
        \caption{(Ideal model) Red arrows depict directions to minimize $\ell_2$ distance between the intermediate images of two reverse paths, $p_1$ and $p_2$. $\mathcal{L}$: loss function.}% Dataset: CIFAR-10.}
        \label{DPC testing}
    %\vspace{-20pt}
    \end{minipage}%   
    \hspace{2pt}
    \begin{minipage}[t]{.45\textwidth}
    \centering
      \centering   
        \includegraphics[width=0.8\linewidth]{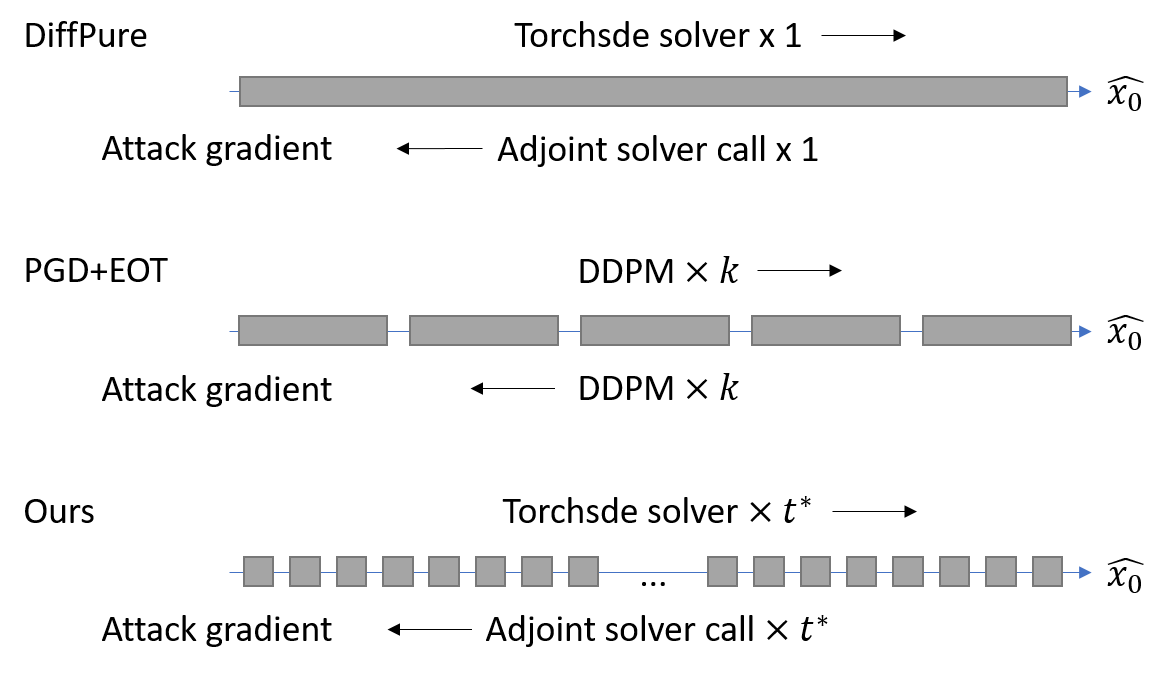}
        \caption{\small Reverse diffusion process implementations: The original implementation of DiffPure involves only one function call in reverse and adjoint solver calls. 
        The PGD+EOT attack utilizes a surrogate diffusion process with fewer steps than purification steps. However, in our implementation, we use the same number of steps for purification and attack.}
        \label{reverse implementation}
    \end{minipage}%
   
\end{figure}

In this test, we verify the framework composed of two diffusion paths, denoted as $p_1$ and $p_2$, and a pre-trained classifier $f_\p$ ({\em e.g.}, pre-trained WRN-28-10), as shown in Fig. \ref{DPC testing}. 
The adaptive adversarial image $x_{adv}$ is generated via BPDA+EOT \cite{Athalye2018} as an input to path $p_1$ while the clean image $x$ is assumed to be available (ideal case) in path $p_2$.
%and used to attack the model composed of purifier plus pre-trained classifier $f_\p$ ({\em e.g.}, pre-trained on WRN-28-10).} 
%The path $p_2$ {\color{blue}first swapped the input} with a ground truth image, aka clean image $x$. 
%The inputs to path $p_1$ and $p_2$ are the adversarial image $x_{adv}$ and its corresponding clean image $x$, respectively.
In this case, we minimize the $\ell_2$ distance between the intermediate image in the reverse process $p_1$ and that in $p_2$, which gives a direction to make $p_1$ close to $p_2$. Finally, the output $\widehat{x^{p_1}}$ is feed into the classifier $f_\p$ for prediction. 
We obtain the natural acc of $93.5\%$ and robust acc of $93.0\%$ from CIFAR-10.
This provides us a hint that the diffusion path $p_1$ should be maintained relatively clean ({\em e.g.}, both the input and reverse diffusion process in path $p_1$ are as clean as those in path $p_2$) so that the output of $p_1$, which is the recovered image $\widehat{x^{p_1}}$, is purified enough.

Therefore, the motivation here is to expand the idea of the opposite adversarial direction in modifying (i) the input for arriving at a safer area and (ii) the entire path for purification.
Nevertheless, the clean image $x$ corresponding to $x_{adv}$ required for the second path $p_2$ is absent during testing. In addition, it is known that adversarial perturbation is added to an image and causes imperceptible changes.
%even if we plot the RGB value of all pixels of an image in three-dimensional space. 
In view of this, we resort to generating purified images as input to $p_2$ using the new reference point strategy, as described in Eq.\! (\ref{eq: data generation rule}) of Sec. \ref{opposite adv}.
%and/or use a process to weaken the adversarial noise.} 

Conceptually, the idea of generating the input to path $p_2$ that guides path $p_1$ is to transfer pixel values from the source image (adversarial image) to the other target image (clean/purified image), which can be treated as finding the optimal transport plan that moves every 3D point (RGB value) in a source point cloud to a target point cloud with the minimum cost ({\em e.g.}, in terms of $\ell_2$ distance between two point clouds).
%%%See Fig. \ref{point cloud and sinkhorn}.
Fortunately, we can use non-attack images, which are the training data, combined with Eq.\! (\ref{eq: data generation rule}) to produce excessively denoised target images for diluting the attack perturbation.

Based on the above test and observations, we now describe the proposed method for cleaning the diffusion path with adversarial images as input.
The flowchart is illustrated in Fig. \ref{defense flow chart}(c).
First, suppose we have $x_{adv}$ as the source image, it will be processed by color transfer with optimal transport \cite{feydy2019interpolating}, which is denoted as ``Color OT'' in Fig. \ref{defense flow chart}(c), using the images coming from the training dataset. 
To this end, we pick $C$ images with one image per class, where $C$ stands for the number of classes. 
By using Eq.\! (\ref{eq: data generation rule}) to generate new reference points from these picked images, we have the target images $x_1^{tar},\ldots,x_C^{tar}$ for ``Color OT'' to change/purify the adversarial pixels in $x_{adv}$. 
The $C$ target images will not be picked again throughout the testing so that there is no randomness.
%Some transferred results are shown in the right part of Fig. \ref{point cloud and sinkhorn}. 

Second, after finding the $x_j^{tar}$ that has the lowest Sinkhorn divergences $S_\ve$ (Eq.\! (3) in \cite{feydy2019interpolating}) with $x_{adv}$, we then use color transfer $f_{CT}$ to modify $x_{adv}$ with reference to $x_j^{tar}$. The output is denoted as $x^{p_2}$. 
The purification procedure is specified as:
% After the {\color{blue}Color OT}, we choose the transformed image, denoted as {\color{blue}$x^{p_2}$}, that corresponds to the lowest {\color{blue}Sinkhorn divergences} {\color{blue} (Eq. (3) in \cite{feydy2019interpolating}),} denoted as $S_\e$:
\begin{equation}
\label{eq: color ot}
    j=\argmin_{i\in\{1,\ldots,C\}}S_\ve(x_{adv},x_i^{tar});\;\;\; %\and\;
    %,\nonumber\\
    x^{p_2}=f_{CT}(x_{adv},x_j^{tar}).
\end{equation}
As our starting point, $x^{p_2}$ goes into the diffusion process, as shown in Fig. \ref{defense flow chart}(c).
This ensures all pixel values in $x^{p_2}$ are not from $x_{adv}$. 
To make it clear, examples of the intermediate images generated from the diffusion process in Fig. \ref{defense flow chart}(c) are illustrated in Fig. \ref{Intermediate Images in 3(c)} of Sec. \ref{Sec: Intermediate Images} in the Supplementary.

%\begin{figure}[h]
%    \centering
%    \includegraphics[width=6cm]{figs/point_cloud_sinkhorn.png}
%    \caption{Left: CIFAR-10 images. Middle: Corresponding 3D point cloud, where each point denotes an RGB pixel value in $[0, 1]$. Right: Results at different iterations (it) of optimal transport.}
%    \label{point cloud and sinkhorn}
%\end{figure}
% \begin{figure}[h]
%     \centering
%     \includegraphics[width=5cm]{figs/sinkhorn.png}
%     \caption{Results of color transfer at each iteration (it) of optimal transport problem.}
%     \label{sinkhorn}
% \end{figure}

Third, we put $x_{adv}$ and $x^{p_2}$ into the diffusion model and set $t^*$, which is the optimal time step \cite{nie2022diffusion} to remove the adversarial noise. 
We maintain two paths: the path with superscript $p_1$ for denoising the color values and the path with superscript $p_2$ for recovering the image.
%Please see the upper and lower paths in Fig. \ref{defense flow chart}(c)}.
%\sout{Through the forward process, $x_{adv}$ and $x^{p_2}$ become $x^{p_1}_{t^*}$ and $x^{p_2}_{t^*}$, respectively.} 
Unlike the test in Fig. \ref{DPC testing}, during the reverse diffusion process, we do not use $p_2$ to pull $p_1$, since $x^{p_2}$ is generated by $f_{CT}$. Instead, we use ``baseline purifier+reverse diffusion'' described in Sec. \ref{Conditional RDP} on one path, $p_1$.
Therefore, after the reverse diffusion process, the image $\widehat{x^{p_2}}$ will refer to the denoised image $\widehat{x^{p_1}}$ as the target for $f_{CT}$ to restore the colors, which is denoted as $\widehat{x_{clean}}$. 
This is because $\widehat{x^{p_2}}$ is still a color-transferred image after the diffusion model, but the output from $p_2$ in the aforementioned test (Fig. \ref{DPC testing}) starts from the ideal clean image $x$ without needing color restoration. The whole process will be iterated again with starting point $\widehat{x_{clean}}$ and $t^*$ being halved at each iteration.
Please see Algorithm \ref{algo:DPC} in Sec. \ref{Sec: algorithm} of Supplementary in describing the entire procedure.

\subsection{Granularity of Gradient Approximation in Realizing Powerful Adaptive AutoAttack}\label{Sec: GGA}
We present to implement a more powerful adaptive AutoAttack via granularity of gradient approximation in order not to overestimate robustness.
Actually, our implementation requires the output in each step from \code{torchsde} in the diffusion reverse process, which starts from $x_{t^*}$ and calls \code{torchsde} to produce the output $x_{t^*-1}$ of next time step till we get the final image $\hat x$. 
Hence, if one understands the mechanism of using \textit{adjoint method} as BPDA correctly, gradient computation in the reverse diffusion process will demand the same amount of calls of \textit{adjoint method} as in that of \code{torchsde}. 
We have to particularly point out that this is different from DiffPure \cite{nie2022diffusion}, where the authors only used one \code{torchsde} call for the final image and one call of \textit{adjoint method} for computing the gradient. 
We believe the granularity (one call vs. multiple calls of \textit{adjoint method}) of gradient approximation causes the performance difference, and the use of multiple calls indeed provides AutoAttack with sufficient information to generate a more powerful adversarial perturbation.
% \sout{Our results in Sec. \ref{Adv AutoAttack-Realization} have validated this point.}

% In addition to Table \ref{Table: Robustness Evaluation}, robustness evaluation in terms of {\AA} (Rand) is further separately treated in  Table \ref{tab: AutoAttack diff} that indicates the difference between our method and DiffPure at first sight.
%We explore the reason behind this surprising result and uncover the insufficiency of realizing {\AA} (Rand) in previous diffusion-based purifiers.
%We speculate that it is probably because our method extracts the output vector $x_t$ from \href{https://github.com/google-research/torchsde}{\code{torchsde}} at each time step of the reverse process and such additional information allows the attacker to estimate the gradient vector more precisely so as to generate adversarial examples more effectively.

To verify our finding, we have observations across different datasets in Table \ref{tab: AutoAttack diff}. First, we selected a subset from CIFAR-10 testing dataset consisting of $64$ images, then generated the corresponding adversarial examples from adaptive AutoAttack (Rand) with $20$ EOT via two different implementations, including (1) AutoAttack (Rand-DiffPure): Original code from DiffPure \cite{nie2022diffusion} using one \code{torchsde} function call and (2) AutoAttack (Rand-Ours): Our own implementation that pulls the output $x_t$ at each time step from \code{torchsde} solver, which means $100$ \code{torchsde} function calls. See Fig. \ref{reverse implementation} for comparison of different implementations of reverse diffusion process.
%There are two kinds of implementations: (1) Original code from DiffPure \cite{nie2022diffusion} using one \code{torchsde} function call and (2) Our own implementation that pulls the output $x_t$ at each time step from \code{torchsde} solver, which means $100$ \code{torchsde} function calls. 
%Table \ref{tab: AutoAttack diff} shows the robustness evaluation on different implementations of diffusion purification, especially the reverse process.
% We transfer the adversarial examples from adaptive AutoAttack (rand) with $20$ EOT attacking purifier+classifier, where the reverse process of purifiers were implemented in two different ways: (1) Original code from DiffPure \cite{nie2022diffusion} using one \code{torchsde} function call and (2) Our implementation that pulls the output $x_t$ at each time step from \code{torchsde} solver, which means $100$ \code{torchsde} function calls. 
%can be added back if room is available: Different from the one described in Sec. \ref{opposite adv}, here we do not use DISCO to update $x_t$ in our implementation since the purpose here is to verify the robustness with a focus only on different implementations of the reverse process.

We can see from Table \ref{tab: AutoAttack diff} that in comparison with AutoAttack (Rand-DiffPure), the defense capability of DiffPure is remarkably reduced (the accuracy in boldface) when the adversarial examples generated from AutoAttack (Rand-Ours) are present, obviously indicating robustness overestimation.
Actually, it is evidence of revealing that our implementation can let attackers create stronger adversarial examples and can be used as a proxy to attack diffusion-based purifiers. Also, this finding sheds light on whether using \textit{adjoint method} hides the information used for creating stronger adversarial examples in an adaptive {\AA} setting. 
 
%if adversarial examples were generated from DiffPure+WRN-28-10 with the original code of DiffPure, our robust accuracy is slightly better than DiffPure. Surprisingly, by using the adversarial examples generated from the same model with our implementation, both the robust accuracy drops remarkably. Our method, however, is demonstrated to be obviously better than DiffPure.
%under our implementation of {\AA} (Rand).
%We are aware that there is a similar discovery in \cite{lee2023robust} but the difference is that the authors employed \textit{surrogate gradient} to approximate the gradient of the reverse process in diffusion purification. \textcolor{blue}{Besides, the attack we discover exhibits transferability among purifiers with different implementations, and it is easy to realize without additional setup of the objective function.}

Finally, since DiffPure \cite{nie2022diffusion} has not been evaluated in \cite{croce2022evaluating}, it is believed that this simple trick of implementation that creates stronger {\AA} (Rand) can be an easy way of attacking test-time adversarial defense purifiers and a promising supplement to \cite{croce2022evaluating}.
In the following experimental evaluations, this kind of adjoint strategy will be used in implementing stronger adaptive attacks.
%, including BPDA, PGD, AutoAttack, and DiffAttack.

\begin{table*}[ht]
  \centering
  \normalsize
  \renewcommand\arraystretch{1.3}
  \resizebox{1.0\textwidth}{!}{
  \begin{tabular}{@{}ccc@{}}
    \toprule
    AutoAttack & (Rand-DiffPure) & (Rand-Ours) \\\hline
    \midrule
    DiffPure & 76.56\% & \textbf{64.06\%}  \\
    %{\color{blue}Sec. \ref{Sec: GGA}} & 78.12\%  & 68.75\%  \\
    \bottomrule
  \end{tabular}\hspace{0.075in}
  \begin{tabular}{@{}ccc@{}}
    \toprule
    AutoAttack & (Rand-DiffPure) & (Rand-Ours)\\\hline
    \midrule
    DiffPure & 26.56\% & \textbf{20.31\%}  \\
    %{\color{blue}Sec. \ref{Sec: GGA}} & 26.56\%  & 25.00\%  \\
    \bottomrule
  \end{tabular}\hspace{0.075in}
  \begin{tabular}{@{}ccc@{}}
    \toprule
    AutoAttack & (Rand-DiffPure) & (Rand-Ours)\\\hline
    %Defenses & Adv from DiffPure & \textcolor{blue}{Adv from Sec. \ref{Sec: GGA}}\\
    \midrule
    DiffPure & 46.88\% & \textbf{28.13\%}  \\
    %{\color{blue}Sec. \ref{Sec: GGA}} & 43.75\%  & 28.125\%  \\
    \bottomrule
   \end{tabular}}
  %\vspace{5pt}
  \caption{\small Robust accuracy for adversarial examples (Adv) generated from different implementations of diffusion purification under adaptive {\AA}  (Rand) with $20$ EOT. Our implementation uses output in every time step from \code{torchsde}, whereas DiffPure \cite{nie2022diffusion} uses \code{torchsde} without accessing the intermediate outputs, which is encapsulated in \code{torchsde} function call. %Adv denotes adversarial sample obtained from adaptive {\AA}  (Rand) with $20$ EOT. 
  Left: CIFAR-10/WRN-28-10; Middle: CIFAR-100/WRN-28-10; Right: ImageNet/ResNet-18.} %Dataset: Subset of CIFAR-10.}
  \label{tab: AutoAttack diff}
\end{table*}

\begin{table*}[ht]
  \centering
  \resizebox{0.9\textwidth}{!}{
  \begin{tabular}{@{}cccc@{}}
    \toprule
    Defense Methods & Clean Accuracy (\%) & Robust Accuracy (\%) & Attacks\\
    \midrule[2pt]
    No defense & 94.78 & 0 & PGD-$\ell_\infty$\\
    \hdashline
    AWP \cite{wu2020adversarial}* & 88.25 & 60.05 & {\AA} (Standard)\\ 
    Anti-Adv \cite{alfarra2022combating}* + AWP \cite{wu2020adversarial} &88.25 & 79.21 & {\AA} (Standard)\\ \hdashline
    DISCO \cite{ho2022disco}*& 89.26 & 82.99 & PGD-$\ell_\infty$\\
    DISCO \cite{ho2022disco} \textbf{+ our OAP ($K=1$)}& \textbf{92.5±2.06} & 88.29±3.3 & PGD-$\ell_\infty$\\
    %Convex comb.\! of DiffPure \cite{nie2022diffusion} and DISCO \cite{ho2022disco} \textcolor{blue}{+ OAG ($K=1$)}& 91.67±2.09 & 89.31±1.96 & PGD-$\ell_\infty$\\
    % DISCO \cite{ho2022disco}*& 89.26 & 82.99 & \multirow{8}{PGD-$\ell_\infty$} &\\
    DiffPure \cite{nie2022diffusion} & 88.06±2.65 & 87.21±2.28 & PGD-$\ell_\infty$\\
    DiffPure \cite{nie2022diffusion} & 88.15±2.86 & 87.71±2.12 & {\AA} (Standard)\\
        %DiffPure \cite{nie2022diffusion} & 89.78±3.91 & 88.28±5.29 & PGD-$\ell_\infty$\\
    %DiffPure \cite{nie2022diffusion} & 89.58±4.29 & \textbf{89.45±5.22} & {\AA} (Standard)\\
    SOAP \cite{shi2021online}*& 96.93 & 63.10 & PGD-$\ell_\infty$\\
    Hill {\em et al}. \cite{hill2020stochastic}*& 84.12 & 78.91 & PGD-$\ell_\infty$\\
    ADP ($\s=0.1$) \cite{yoon2021adversarial}*& 93.09 & 85.45 & PGD-$\ell_\infty$\\
%    GDMP (SSIM) \cite{wang2022guided,wu2022guided}& 93.50 & 90.10 && \\
    % Ours ((\ref{eq: data generation rule}) with $K=1$+DISCO) &  & & & \\
    % Ours ((\ref{eq: data generation rule}) with $K=1$+DiffPure) &  & & & \\
    % Ours ((\ref{eq: data generation rule}) with $K=3$+DiffPure) &  &  & & \\
    Ours (Sec. \ref{Conditional RDP}) %with $\eta=$ 5e-3) 
    & 90.77±2.25 & \textbf{88.48±2.04} & PGD-$\ell_\infty$\\
    % Ours (Sec. \ref{DPC} with $\eta=$ 2.5e-3) & 88.93±5.47 & 88.22±4.74 & N/A & PGD-$\ell_\infty$\\ 
    % \hdashline
    Ours (Sec. \ref{Conditional RDP}) %with $\eta=$ 5e-3) 
    & \textbf{90.46±2.36} & \textbf{89.06±2.62} & {\AA} (Standard)\\
    % Ours (Sec. \ref{DPC} with $\eta=$ 2.5e-3) & 89.29±2.13 & 88.17±1.89 & N/A & {\AA} (Standard)\\
    \bottomrule
  \end{tabular}}
  \caption{\small Non-adaptive robustness comparison between our method and state-of-the-art methods.
%  , including Anti-Adv \cite{alfarra2022combating}, DISCO \cite{ho2022disco}, DiffPure \cite{nie2022diffusion}, and GDMP \cite{wang2022guided,wu2022guided}. 
 % All comparisons were conducted under the same conditions (the same attack and the same NN model).
 Classifier: WRN-28-10.
 Asterisk (*) indicates that the results were excerpted from the papers. Boldface indicates the best performance for each attack.
Note that, by incorporating our \textit{Opposite Adversarial Path} (OAP) prior, the clean and robust accuracy of DISCO can be greatly increased.}
  \label{Table: Robustness Evaluation Non-adaptive}
  % \vspace{-10pt}
\end{table*}

\begin{table*}[ht]
% \vspace{-10pt}
  \centering
  \resizebox{0.75\textwidth}{!}{
  \begin{tabular}{@{}cccrc@{}}
    \toprule
    Defense Methods & Clean Accuracy (\%) & Robust Accuracy (\%) & Attack time cost (sec.) & Attacks\\
    \midrule[2pt]
    % No defense & 95.18 & 0 & BPDA+EOT& WRN-28-10 \\ 
    % No defense & 95.18 & 0 & \multirow{7}{*}{BPDA+EOT}& WRN-28-10 \\ 
    % \hdashline DISCO \cite{ho2022disco}&  89.12 & 47.18 & BPDA only & VGG16\\ \hdashline
    No defense & 94.78 & 0 & N/A & BPDA+EOT\\
    DiffPure &  \textbf{92.38±1.86} & 80.92±3.53& 592.92 & BPDA+EOT\\
%    GDMP \cite{wang2022guided,wu2022guided}*& 93.50 & 79.83&& WRN-28-10\\
    Hill {\em et al.}* & 84.12 & 54.90 & N/A & BPDA+EOT\\
    ADP ($\s=0.1$)*& 86.14 & 70.01 & N/A & BPDA+EOT\\
%    Ours (Sec. \ref{Conditional RDP} with $\eta=$2.5e-3) & 92.33±1.48 & 80.23±2.17 & BPDA+EOT& ResNet18\\
    % Ours (Sec. \ref{Conditional RDP} with $\eta=$ 2.5e-3) & 92.50±1.62 & 81.10±2.47 & -- & BPDA+EOT\\
    Ours (Sec. \ref{DPC}) %with $\eta=$ 2.5e-3) 
    & 92.08±1.99 & \textbf{81.25±3.62} & \textbf{6880.97} & BPDA+EOT\\ \hdashline
    %\midrule[2pt]
    %Def1 & 0 & 0 & &\\
    %Def2 & 0 & 0 & &\\
    %Def3 & 0 & 0 & &\\
    %Def4 & 0 & 0 & &\\
%    DiffPure & \textbf{82.81} & 64.06 & 6321.96 &AutoAttack (Rand-DiffPure)\\
%    DiffPure & 81.25 & \textit{45.31} & {\color{blue}11279.92} & \textbf{AutoAttack (Rand-Ours)}\\ 
%    Ours (Sec. \ref{DPC}) %with $\eta=$ 1e-3)
%    & \textbf{82.81} & {\color{blue}--} & {\color{blue}--} & \textbf{AutoAttack (Rand-Ours)}\\ \hdashline
    DiffPure& 96.88 & 46.88 & 3632.94 &PGD+EOT\\
    % Ours (Sec. \ref{Conditional RDP} with $\eta=$ 2.5e-3)&&48.96 & -- & PGD+EOT\\
    Ours (Sec. \ref{DPC}) %with $\eta=$ 2.5e-3)
    & \textbf{100} & \textbf{53.12} & \textbf{22721.90} & PGD+EOT\\ \hdashline
    % Ours (Sec. \ref{Conditional RDP} with $\eta=$ 2.5e-3)&&& -- & DiffAttack\\
    DiffPure &89.02&46.88& N/A &DiffAttack\\
%    Ours (Sec. \ref{Conditional RDP} with $\eta=$ 2.5e-3)&{\color{blue}96.88}&{\color{blue}62.50}& {\color{blue}11570.46} &DiffAttack\\
    Ours (Sec. \ref{DPC}) %with $\eta=$ 2.5e-3)
    & \textbf{95.31}& \textbf{93.75}& \textbf{20397.27} &DiffAttack\\
    \bottomrule
  \end{tabular}}
  \caption{\small Aadaptive robustness comparison between our method and state-of-the-art methods with attack time cost per image.
%  , including Anti-Adv \cite{alfarra2022combating}, DISCO \cite{ho2022disco}, DiffPure \cite{nie2022diffusion}, and GDMP \cite{wang2022guided,wu2022guided}. 
 % All comparisons were conducted under the same conditions (the same attack and the same NN model).
 Classifier: WRN-28-10.
 Asterisk (*) indicates that the results were excerpted from the papers. Boldface indicates the best performance for each attack.
 The attacks include BPDA+EOT, PGD+EOT \cite{lee2023robust}, and DiffAttack \cite{kang2024diffattack}.}
  \label{Table: Robustness Evaluation Adaptive}
\end{table*}

\subsection{Attack Cost and Time Complexity}\label{Attack Cost}
We study how to resist adaptive attacks by analyzing and increasing the time cost of breaking the proposed defense models.
The results are shown in Table \ref{Table: Robustness Evaluation Adaptive}. 
Due to space constraints, please see the time complexity analysis in Sec. \ref{Adaptive Attacks} in the Supplementary for details.

%{\color{blue} reduced in half (e.g. $t^*=100$ for the first iteration, and in the second iteration $t^*$ reduced to 50)}. 
%See Fig. \ref{defense flow chart}.
\section{Experiments}
We examine the performance of proposed test-time adversarial defense methods, described in Sec. \ref{Conditional RDP} and Sec. \ref{DPC}.
%, against state-of-the-art adversarial attacks, and performance comparison with SOTA purification-based defenses.
% Then, we exemplify the pitfall of conducting AutoAttack (Rand) for diffusion-based defense methods. 
%The goal is to verify whether the proposed purifiers can improve the robustness of an existing defense mechanism if our purifier is plugged in.
%we test our method under several attacks. Since our method includes a diffusion model, we conduct our experiments under the scheme of DiffPure \cite{nie2022diffusion} and other related works. The experiments are in a non-adaptive and adaptive attack manner. We also test our method in the BPDA+EOT scheme to overcome randomness in the diffusion model.

\subsection{Datasets and Experimental Settings}\label{datasets_exp_settings}

Three datasets, CIFAR-10 \cite{krizhevsky2009learning}, CIFAR-100 \cite{krizhevsky2009learning}, and ImageNet \cite{deng2009imagenet}, were adopted, where the results for CIFAR-100 and ImageNet are shown in Table \ref{tab: AutoAttack diff} and Sec. \ref{Sec: More Results} of Supplementary.
All experiments were conducted on a server with Intel Xeon(R) Platinum 8280 CPU and NVIDIA V100.

For a fair comparison, we followed RobustBench \cite{Robustbench2021} and existing literature to conduct experiments on two popular NN models, including ResNet-18 \cite{he2016deep} and WRN-28-10 \cite{WRN}.
The step size, $\eta$, in Eq.\! (\ref{Eq: Double Directions}) of Sec. \ref{Conditional RDP} was set as $2.5\times10^{-3}$ and we followed \cite{nie2022diffusion} to set $t^*$ used in Sec. \ref{Conditional RDP} and Sec. \ref{DPC} as $0.1$. Since $t^*=0.1$, the number of steps required in the reverse process is $100$, where the step size $dt$ for \code{torchsde} solver is set to 1e-3.
%For diffusion path cleaning-based purifier in Sec. \ref{DPC}, 
We set $\ve=0.05$ in Eq.\! (\ref{eq: color ot}), which is the default setting in the official package (\href{https://www.kernel-operations.io/geomloss/index.html}{GeomLoss}) \cite{feydy2019interpolating}. For all attacks, we used $\ell_\infty$ and set perturbation   
%allowed perturbation $\d$ such that 
$\norm{\delta}_\infty\le8/255$.

%\subsection{Datasets for Training and Testing}
 %First, we verify the idea of opposite directions on non-adaptive PGD and AutoAttack, and compare with others. 
For training, the only model that needs to be trained is the baseline purifier $g_{\theta_n}$ with $K=1$, which we chose DISCO \cite{ho2022disco} as the baseline to be combined with our new reference point generation in Eq.\! (\ref{eq: data generation rule}) with $K=1$ throughout the experiments. 
In computing the attack gradient per step ($K$), we used PGD-$\ell_\infty$ with 7 iterations. 
%The training procedure followed the setup in \cite{ho2022disco}.
For testing the diffusion-based purifiers, we followed the testing paradigm described in  \href{https://github.com/NVlabs/DiffPure}{DiffPure} \cite{nie2022diffusion}, including the uses of $24$ random subsets (each contains $64$ images) for AutoAttack and $15$ random subsets (each contains $200$ images) for BPDA+EOT from CIFAR-10 testing dataset. 
%{\color{blue} In addition, for non-adaptive cases, we followed the subset setting of AutoAttack.}
%, respectively given by 24 random seeds.}

\subsection{Adversarial Robustness Evaluations}\label{Adv Robust Eval}
Several types of adversarial attacks, including (A1) non-adaptive attacks (PGD-$\ell_\infty$ \cite{madry2017towards}, AutoAttack (Standard) \cite{AutoAttack2020}), (A2) adaptive attacks (BPDA+EOT \cite{Athalye2018}, PGD+EOT \cite{lee2023robust} and DiffAttack \cite{kang2024diffattack}), and (A3) $\ell_2$-norm optimization-based attacks and black-box attacks, including C\&W attack \cite{carlini2016towards}, SPSA \cite{uesato2018adversarial}, and the targeted black-box attack in both settings (nonadaptive/adaptive),
were adopted. Please see Sec. \ref{Sec: More Results} of Supplementary for the evaluation of (A3).

For AutoAttack, we utilized the package \href{https://github.com/fra31/auto-attack}{AutoAttack} \cite{AutoAttack2020} with $\ell_\infty$, in which it has two settings: (1) ``Standard,'' which includes APGD-CE, APGD-DLR, FAB, and Square Attack and (2) ``Rand,'' which includes APGD-CE and APGD-DLR with Expectation Over Time (EOT) \cite{Athalye2018} in case of models with stochastic components. To the most extreme case in which the attacker knows every detail about our framework of ``purifier+classifier,'' we utilized BPDA (\textit{adjoint method} \cite{li2020scalable}) 
%for the diffusion model) 
to bypass purifiers and EOT to combat the randomness in purifiers.
As mentioned in Sec. \ref{Sec: GGA}, our adjoint strategy will be used to implement stronger adaptive attacks in order to avoid robustness overestimation.

Robustness performance was measured by clean/natural accuracy (Clean Acc) for benign samples and robust accuracy (Robust Acc) for adversarial samples.
Several test-time adversarial defense methods, including Anti-Adv \cite{alfarra2022combating}, DISCO \cite{ho2022disco}, DiffPure \cite{nie2022diffusion}, SOAP \cite{shi2021online}, Hill {\em et al}. \cite{hill2020stochastic}, and ADP \cite{yoon2021adversarial}, were adopted for comparison.
%and GDMP \cite{wang2022guided,wu2022guided}, were adopted.
Tables \ref{Table: Robustness Evaluation Non-adaptive} and  \ref{Table: Robustness Evaluation Adaptive} show the robustness evaluations and indicate that our methods either outperform or are comparable with prior works. 
%The results of Anti-Adv \cite{alfarra2022combating}, DISCO \cite{ho2022disco}, Diffpure \cite{nie2022diffusion}, and GDMP \cite{wang2022guided,wu2022guided} are excerpted from the papers.

The experiment in Table \ref{Table: Robustness Evaluation Non-adaptive} is under the setting of non-adaptive attacks (PGD-$\ell_\infty$ with 40 iterations and AutoAttack (Standard)), in which the attacker only knows the information of the downstream classifier. According to \cite{alfarra2022combating}, we specifically point out that the authors used the robustly trained classifier, Adversarial Weight Perturbation (AWP) \cite{wu2020adversarial}, as the testing classifier.
So, except \cite{alfarra2022combating}, we used a normally trained classifier throughout the experiments.

Table \ref{Table: Robustness Evaluation Adaptive} shows the results obtained under adaptive attacks, including stronger ones like PGD+EOT  \cite{lee2023robust} and DiffAttack \cite{kang2024diffattack}.
For the two kinds of AutoAttack (Rand) described in Sec. \ref{Sec: GGA}, please refer to Table \ref{tab: AutoAttack diff}.
Since most diffusion-based purifiers exhibit randomness, %and the study in \cite{nie2022diffusion} reveals that they experience lower robust accuracy under \href{https://github.com/fra31/auto-attack}{AutoAttack} \cite{AutoAttack2020} with random (EOT attack \cite{Athalye2018}) setting. 
we utilized the ``EOT'' setting for randomness, and ``BPDA'' for bypassing the reverse process of diffusion-based methods, which use the \textit{adjoint method} to calculate the gradient of such process. 
%\textcolor{red}{Moreover, according to Sec. \ref{Sec: GGA}, there are two implementations: (1) The first is to generate adversarial examples by DiffPure; (2) The second is to generate adversarial examples by our method.} 
%Our evaluation was also conducted using stronger adaptive attacks \cite{lee2023robust}, \cite{kang2024diffattack}. 
We also provide the time needed to attack an image (attack time cost) against a defense method.
Besides, according to the dual-paths design of Sec. \ref{DPC}, all adaptive attacks have to attack both paths. As a result, our defense experiences TWICE stronger attacks than other single-path methods since gradients are obtained from two paths. In other words, attacks are computed twice.
%\subsubsection{Non-adaptive attack}
%\subsubsection{AutoAttack}
%\subsubsection{Adaptive Attack}

%To summarize the results in Tables \ref{Table: Robustness Evaluation Non-adaptive} and  \ref{Table: Robustness Evaluation Adaptive}, our methods (Sec. \ref{Conditional RDP} and Sec. \ref{DPC}) either outperform or are comparable with the prior works.
%In addition, incorporating our \textit{Opposite Adversarial Gradient} (OAG) prior, the clean and robust accuracy of DISCO can be greatly increased. %On the other hand, diffusion-based purifier provides baseline performance for robust accuracy (e.g.\! 88.28\%), which we suspect is due to the powerful prior data distribution that models possess.}
More specific, we can see from Table \ref{Table: Robustness Evaluation Adaptive} that, in addition to accuracy, the time costs  the attackers need to generate attack examples for our defense method are greatly higher than those for other defense methods. 
If the attackers would like to shorten computations of generating adversarial examples,  the number of iterations of conducting attacks or the number of EOT need to be reduced, thereby weakening the attack performance. 
%On the other hand, if attackers need a stronger attack, they have to experience more computational burden. 
Take BPDA+EOT as an example: the total time to finish a batch testing on DiffPure costs less than 1 day but it costs 2 days to test on our proposed method under the same setting with 8 V100 GPUs. 
Moreover, the number of paths in our method (Sec. \ref{DPC}) can be flexibly increased to be larger than two to greatly increase the time cost for attackers to generate adaptive attack examples.
An accompanying merit is that the robust accuracy of our method in resisting DiffAttack is rather high because  %Current diffusion-based purifiers only use a single path to clean the image. Hence, 
DiffAttack focuses on attacking the only one path by computing the gradient on it without meeting our dual path strategy. %gradients from two different paths might confuse such an attack, which leads to better performance.}

\section{Conclusions \& Limitations}\label{section-conclusions}
We have presented a new test-time adversarial defense method by 
%The key 
%to adversarial robustness 
%is to 
excessively denoising the incoming input image %(benign or adversarial) 
along the opposite adversarial path (OAP). % so as to move far away from the decision boundary.
This OAP prior can be readily plugged into the existing defense mechanisms for robustness improvement. 
%Our defense method also forces attackers to spend a great deal of time creating adaptive adversarial examples.
Meanwhile, we exemplify, for the first time, the pitfall of conducting {\AA} (Rand) for diffusion-based adversarial defense methods. 
However, we are aware there are several attacks targeting diffusion-based adversarial defenses, and the performance of our proposed method may be overestimated since the gradient computation is approximated.
%%%\textcolor{blue}{Besides, such defense does not exhibit transferability when applied to other datasets, it requires training of diffusion model and OAG purifiers, which causes large training overhead.}
%Experimental verification with state-of-art attacks demonstrates the effectiveness our methods.

\clearpage
\bibliographystyle{unsrtnat}
\bibliography{template}

\clearpage
\newpage
\appendix
\onecolumn

\section*{Appendix}

\section{Related Works: Supplement}\label{Related Works: Supp}
In \cite{hill2020stochastic}, the author proposed incorporating an energy-based model (EBM) with Markov Chain Monte Carlo (MCMC) sampling for adversarial purification.
This method constitutes a memoryless sampling trajectory that removes adversarial signals, while this sampling behavior preserves image classes over long-run trajectories.

In Adaptive Denoising Purification
 (ADP) \cite{yoon2021adversarial}, the authors used the Noise Conditional Score Network (NCSN) with Denoising Score Matching (DSM) as the purifier, but with a deterministic short-run update rule for purification. This fixes the need for performing long-run sampling in order to remove adversarial noise in \cite{hill2020stochastic}.

Guided Diffusion Model for Purification (GDMP) \cite{wang2022guided, wu2022guided} is proposed to embed purification into the reverse diffusion process of a DDPM \cite{ho2020denoising}. GDMP submerges adversarial perturbations with gradually added Gaussian noises during the diffusion process and removes both noises through a guided denoising process. By doing so, GDMP can significantly reduce the perturbations raised by adversarial attacks and improve the robustness of classification.

\section{Adversarial Attacks}\label{Sec: AA}
Two types of adversarial attacks are briefly introduced here.

{\bf White-box attack.} The attacker knows all information about $f_\p$, including the model architecture, parameters $\p$, training schedule, and so on.
One of the most indicative white-box attacks is projected gradient descent (PGD) \cite{madry2017towards}, a gradient-based attack. It produces {\AP} by projecting NN's gradients on the clipping bound in an iterative manner. 
If the gradient is correctly calculated, the loss would certainly be maximized, and the NN, especially for non-robust NN, will be likely to return misclassified outputs.

{\bf Black-box attack.} The attacker does not know all the information mentioned above in $f_\p$. Common black box attack methods are roughly divided into two types: Query-base attack and Transfer-based attack.

%(1) Query-base attack: through the query method, that is, through the interaction between input and output and $f_\p$, the probability distribution of each category is obtained to estimate $f_\p$ The gradient of the loss function. (2) Transfer-based attack: By training a surrogate NN $\tilde{f}_\p$ that is as similar as possible to $f_\p$, a white-box attack can be performed on $\tilde{f}_\p$.

No matter white- or black-box attack is concerned, in order to make {\AP} less detectable, we will set a range of attack intensity, that is, {\AP} is only allowed to perturb within a given norm value. 
Usually $\ell_p$-norm, denoted as $\norm{\cdot}_p$ ($p=1,2,\infty$), is used:
\begin{align}
    \norm{\delta}_p:=\left\{
    \begin{array}{ll}
        \left(\sum_i\delta_i^p\right)^{1/p} & p=1 \mbox{ or } 2, \\
        \max_i\abs{\delta_i} & p=\infty.
    \end{array}
    \right.
\end{align}
%It can be seen from the above formula that under the same norm, the intensity of $l_\infty$-norm is the largest; 
In terms of intensity, we are accustomed to using $\e_p$ to represent it. For example, it often uses $\e_\infty=8/255$ to indicate that the intensity range of currently used $\delta$ is $\norm{\delta}_\infty\le8/255$.
\section{Diffusion Models: Supplement}\label{Sec: DMs: Supp}
Diffusion models were inspired by the diffusion phenomena under nonequilibrium thermodynamics in the physical world to design a framework that generates data by learning the reverse process of the data being destroyed by Gaussian noise gradually.

In the literature, the diffusion model \cite{sohl2015deep} is a type of generative model ({\em e.g.}, GAN and VAE).
Conceptually, this generative process behaves like denoising. 
Given a data point $x_0\sim q$, where $q$ denotes the (unknown) true data distribution, and a variance schedule $\{\beta_t\}_{t=1}^T$, the forward diffusion process follows $q(x_{1:T}|x_0)=\prod_{t=1}^Tq(x_t|x_{t-1})$,
%    &\mbox{where }q(x_t|x_{t-1})
%    =\cN(x_t;\sqrt{1-\b_t}x_{t-1},\b_tI),\nonumber
where $q(x_t|x_{t-1})=\cN(x_t;\sqrt{1-\beta_t}x_{t-1},\beta_tI)$, and the reverse diffusion process follows:
\begin{align}
    p_\theta(x_{0:T})
    =p(x_T)\prod_{t=1}^Tp_\theta(x_{t-1}|x_t),
\label{Eq: Reverse-Diffusion}
\end{align}
where $p_\theta(x_{t-1}|x_t)\sim\cN(x_{t-1};\m_\theta(x_t,t),\SG_\theta(x_t,t))$,
$x_T\sim\cN(0,I)$, and $\m_\theta(x_t,t)$ and $\SG_\theta(x_t,t)$ denote the mean and covariance from the diffusion model parameterized by $\theta$ at time step $t$, respectively.

After that, there are several types of recently developed diffusion models, including score-based diffusion \cite{song2019generative,song2020score}, guided-diffusion \cite{dhariwal2021diffusion}, ILVR \cite{choi2021ilvr}, denoising diffusion probabilistic model (DDPM) \cite{ho2020denoising}, and DDA \cite{gao2023back}.
%We will describe guided-diffusion here and please refer to Sec. \ref{Sec: DMs: Supp} in the Supplementary for details.

Specifically, in guided diffusion \cite{dhariwal2021diffusion}, given a label $y$ as the condition and Eq.\! (\ref{Eq: Reverse-Diffusion}), the conditional reverse process is specified as:
\begin{align}\label{eq: guided reverse}
    p_\theta(x_0,\ldots,x_{T-1}|x_T,y)
    =\prod_{t=1}^Tp_\theta(x_{t-1}|x_t,y).
\end{align}
% For simplicity(???why), we may consider one term 
To solve 
%the reversed output in 
Eq.\! (\ref{eq: guided reverse}), $\theta$ is decomposed into two terms as $\theta=\vp\cup\p$ to form separate models:
\begin{align} \label{equ condition}
    p_\theta(x_{t-1}|x_t,y)
    =Zp_\vp(x_{t-1}|x_t)p_\p(y|x_{t-1}),
\end{align}
where $Z$ is a normalization constant.
Guided-diffusion improves the model architecture by adding attention head and adaptive group normalization, that is, adding time step and class embedding to each residual block. At the same time, with reference to GAN-based conditional image synthesis, class information is added during sampling and another classifier is used to improve the diffusion generator. To be precise, the pre-trained diffusion model can be adjusted using the gradient of classifier to direct the diffusion sampling process to any label.

Score-based diffusion \cite{song2019generative,song2020score} generates samples by estimating the gradients of unknown data distribution with score matching, followed by Langevin dynamics, moving data points to areas with higher density of data distribution.
% generates samples by Langevin dynamics using gradients of the data distribution estimated with score matching.
In practice, the score network $s_\theta$ is trained to predict the true data distribution $q$ as:
\begin{align}
    s_\theta(x_t,t)
    \approx\nabla_{x_t}\log q(x_t)
    =-\frac{\mathbf{\e_{\theta}}(x_t,t)}{\sqrt {1-\bar{\alpha}_t}},
\end{align}
where $\bar{\a}_t=\prod_{s=1}^t(1-\beta_s)$.

% \begin{align*}
%     p_t(x_{nf}|x_t)
%     =p_t
% \end{align*}

%\subsection{Defense Methods}\label{Defenses}
%{\bf ILVR \cite{choi2021ilvr}, DDA \cite{gao2023back}}. 
On the other hand, the conditional generation of the diffusion model has also received considerable attention. 
In ILVR \cite{choi2021ilvr}, the author proposed a learning-free conditioning generation, which is challenging in denoising diffusion probabilistic model (DDPM) \cite{ho2020denoising} due to the stochasticity of the generative process. It leveraged a linear low-pass filtering operation $\p_N$ as a condition to guide the generative process in DDPM for generating high-quality images based on a given reference image $c$ at time $t$, termed $c_t$, which can be obtained by the forward diffusion process $q$. The update rules are derived as follows:
\begin{align}
    &x'_t \sim p_{\theta}(x'_t|x_{t+1}) \nonumber \\
    &c_t \sim q(c_t|c) \nonumber \\
    &x_t \leftarrow \p_N(c_t) - x'_t  - \p_N(\hat x'_t),
\end{align}
where the factor of downsampling and upsampling is denoted as $N$.

DDA \cite{gao2023back} also came up with a similar approach to resolve the domain adaptation problem in the test-time scenario. The authors adapt the linear low-pass filtering operation $\p_N$ in ILVR \cite{choi2021ilvr} as conditions, and their method also forces the sample $x_t$ to move in the direction that decreases the distance between the low-pass filtered reference image $\p_N(x_0)$ and low-pass filtered estimated reference image $\p_N(\hat x_0)$. The update rule is specified as follows:
\begin{align}
    &\hat x_0 \leftarrow \sqrt{\frac{1}{\Bar{\alpha}_t}\,}\,x_t-\sqrt{\frac{1}{\Bar{\alpha}_t}-1\,}\,\mathbf{\e_{\theta}}(x_t, t) \\
    &x_t \leftarrow \hat x_t - \textit{w}\nabla_{x_t} \norm{\p_N(x_0) - \p_N(\hat x_0)}_2,
\end{align}
where $N$ is the factor of downsampling and upsampling, and $\textit{w}$ is the step size.

% {\bf Pseudoinverse-guided DM} \cite{song2022pseudoinverse} applied conditional guidance to solve various common inverse problems, including super-resolution, inpainting, JPEG reconstruction, etc. They started from giving an observation $y$ and a known measurement $H$, then estimate the gradient of $\log p_t(x_t|y)$, with the form:
% \begin{align*}
%     &\nabla_{x_t}p_t(x_t|y)\\
%     &\quad=\nabla_{x_t}p_t(x_t)+\nabla_{x_t}p_t(y|x_t)\\
%     &\quad\approx\nabla_{x_t}p_t(x_t)+r_t^{-2}\left((H^\dagger y-H^\dagger H\hat{x}_t)^\top\frac{\partial\hat{x}_t}{\partial x_t}\right)^\top
% \end{align*}
% where $H^\dagger=^\top(HH^\top)^{-1}$ is the pseudoinverse of $H$; and $\hat{x}_t$ is a solution of Tweedie's formula.
\section{Intermediate Images Generated from Fig. \ref{defense flow chart}(c)}\label{Sec: Intermediate Images}

\begin{figure*}[t]
    \centering    \includegraphics[width=0.8\textwidth]{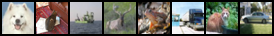}    
    \centering    \includegraphics[width=0.8\textwidth]{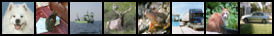}
    \centering    \includegraphics[width=0.8\textwidth]{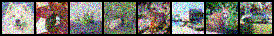}
    \centering    \includegraphics[width=0.8\textwidth]{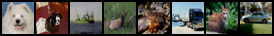}
    \centering    \includegraphics[width=0.8\textwidth]{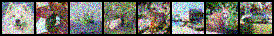}
    \centering    \includegraphics[width=0.8\textwidth]{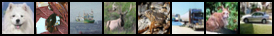}
    \centering    \includegraphics[width=0.8\textwidth]{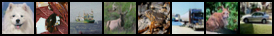}
    \caption{Intermediate Images generated from Fig. \ref{defense flow chart}(c). From Top to Bottom: The images denote clean image $x$, %adversarial image $x_{adv}$, 
    $x^{p_2}$, $x^{p_1}_{t^*}$,
    $\widehat{x^{p_1}}$,
    $x^{p_2}_{t^*}$,  $\widehat{x^{p_2}}$, and purified image $\widehat{x_{clean}}$, respectively.}
    \label{Intermediate Images in 3(c)}
\end{figure*}

In Fig. \ref{Intermediate Images in 3(c)}, we show the images generated from each step in Fig. \ref{defense flow chart}(c) for visual inspection.

% \vspace{0.8cm}
% {\color{blue}
\section{Algorithm in Sec. \ref{DPC}}\label{Sec: algorithm}

Here, we describe the entire procedure of the proposed method in Sec. \ref{DPC}.
%The symbols and functions that appear in Algorithm \ref{algo:DPC} are all mentioned in Sec. \ref{DPC}.
% \begin{algorithm}
% {\color{blue}
% \caption{{\color{blue}Section \ref{DPC}}}
% \begin{algorithmic}[1]
% \REQUIRE Purifier $g_\theta$, adversarial image $x_{adv}$
% \ENSURE Purified image $\widehat{x_{clean}}$
% \STATE $x\leftarrow x_{adv}$
% \FOR{repeated time from $1\ldots2$}
%     % \STATE \textbf{..}
%     \STATE $x^{p_1}\leftarrow x$; $x^{p_2}\leftarrow x$
%     \STATE $j\leftarrow \argmin_{i\in\{1,\ldots,C\}}S_{\ve}(x^{p_2}, x_i^{tar})$
%     \STATE $x^{p_2}\leftarrow f_{CT}(x^{p_2},x_j^{tar})$
%     \STATE $x^{p_1}\leftarrow g_\theta(x^{p_1})$; $x^{p_2}\leftarrow g_\theta(x^{p_2})$
%     % \FOR{time step $t$ in $t^\ast\ldots1$
%     % \ENDFOR
%     % \STATE \textbf{Evaluation Phase}
%     \STATE $x\leftarrow f_{CT}(x^{p_2},x^{p_1})$
% \ENDFOR
% \STATE $\widehat{x_{clean}} \leftarrow x$
% \RETURN $\widehat{x_{clean}}$
% \end{algorithmic}
% \label{algo:DPC}
% }
% \end{algorithm}
% }

% https://tex.stackexchange.com/questions/216096/how-to-color-an-entire-algorithm-environment-in-latex
% \lipsum[1]

% \section{My Algorithm}

% \lipsum[2]

\floatname{algorithm}{\color{black}Algorithm}
\begin{algorithm}
  \caption{\color{black}Diffusion Path Cleaning-based Purifier}
  \color{black}
  \begin{algorithmic}[1]
    \input{\jobname2.alg}
  \end{algorithmic}
\label{algo:DPC}
\end{algorithm}

% \lipsum[3-5]
% }

\section{Attack Cost in Time Complexity: Detailed Analysis}\label{Adaptive Attacks}
In this section, we discuss the cost the attackers need to pay to defeat our proposed test-time adversarial defense method. In particular, we focus on analyzing the time complexity of defeating the diffusion-based purifiers presented in Sec. \ref{Conditional RDP} and Sec. \ref{DPC}.

In DISCO \cite{ho2022disco}, let $N_d$ and $N_c$ be defined as the number of parameters in DISCO and its downstream classifier, respectively. 
In \textit{training time}, the authors estimated that the time complexity of defense/purification is $\cO(N_d)$ and that of adaptive attack ({\em i.e.}, designing an adversarial example by knowing the whole model) is $\cO(KN_d+N_c)$, where $K$ is the number of steps. 
Hence, the ratio of attack-to-defense cost in training time is $\cO(K+N_c/N_d)$ and the authors asserted that the time complexity of defense is much less than that of attack since $N_d<N_c$.
In \textit{testing time}, the time complexity of defense becomes $\cO(KN_d)$ (see green bars in Fig 10 of \cite{ho2022disco}) while the other remains the same. 
Hence, the ratio of attack-to-defense cost in testing time is $\cO(1+N_c/(KN_d))$, revealing that even $N_c$ is larger than $N_d$, the time complexity for both the attacker and defender can tie if $K$ is large enough.

In DiffPure \cite{nie2022diffusion}, the authors argued that an adaptive attack on diffusion model by traditional back-propagation would cause high memory cost.
To overcome this issue, they instead applied {\it adjoint method} \cite{li2020scalable} to efficiently estimate the gradient used for designing adaptive adversarial perturbation under constant memory cost. 
From Table 1 in \cite{li2020scalable}, it is asserted that, under the {\it tolerance $\e=1/T$}, the per-step time complexity scales as $\log_2T$, where $T$ is the number of steps required during the reversed process. So, the time complexity of adjoint method through the reverse diffusion process is $\cO(T\log_2 T)$. 

We, however, raise the concern that the presented time cost is likely to ignore one important factor: the number of parameters in a diffusion model, denoted as $N_{dm}$. 
Hence, in conservatively speaking, the time complexity of adjoint method through the reverse process should be corrected as $\cO(N_{dm}T\log_2 T)$.

% These observations motivate us ???to investigate an adversarial defense method that can significantly increase the cost of attacks in preparing ad adversarial examples, as described in Sec. \ref{Conditional RDP} and Sec. \ref{DPC}.

% Let us consider the case of a single image $x$.
% Let $L=L(\hat{y},\p,\th)$, where $\hat{y}=f_\p(g_\th(x))$. 
% The time complexity of finding the adversarial example $x_{adv}$ is dominated by computing $\pl L/\pl x$. By the chain rule,
% \begin{align*}
%     \frac{\pl L}{\pl x}
%     =\frac{\pl L}{\pl\hat{y}}\frac{\pl\hat{y}}{\pl x}
% \end{align*}
% or
% \begin{align}
%     \frac{\pl L}{\pl x}
%     =\frac{\pl L}{\pl z}\frac{\pl z}{\pl x},
% \end{align}
% where $z$ denotes the output of some hidden layer.
% In particular, if $z=wx+b$, then $\pl z/\pl x=w$, where $w$ denotes layer weights. 
% Note that if a defense follows gradient masking, then $w$ is close to a zero vector, and so does $\pl L/\pl x\approx 0$.

% ???Define $T(c,m,n)$ as the time complexity of $y=Ax$, where $A$ is any matrix of size $c\times m\times n$ and $x$ is of size $c\times n$.

Based on the above concerns, our estimations of the time complexity of adversarial attack and proposed defense methods in {\it testing time} are described as follows.
For the defense method described in Sec. \ref{Conditional RDP},
% we denote the number of steps required for the reverse diffusion process as $T$.
since the baseline purifier $g_\theta$ is reused in every step of the diffusion process, $T$ is equal to $K$ (hereafter, we will use them interchangeably). 
So, the time complexity costs of adaptive attack and our defense are derived as $\cO(N_{dm}T\log_2 T+TN_d+N_c)$ and $O(N_{dm}T+N_dT+N_c)$, respectively.
In practice, $N_{dm}$ is usually much larger than $N_d$ and $N_c$, so we have $\cO(N_{dm}T\log_2 T+TN_d+N_c)\approx\cO(N_{dm}T\log_2 T)$ and $\cO(N_{dm}T+N_dT+N_c)\approx\cO(N_{dm}T)$, and the ratio of attack-to-defense cost is $\cO(\log_2 T)$. Moreover, if the attacker adopts the Expectation Over Time (EOT) operation with a number of iterations, $T_{EOT}$, the time cost of creating such an attack has to be additionally multiplied by $T_{EOT}$. 
For example, if $T=100$ and $T_{EOT}=20$, the ratio theoretically approximates $133$; {\em i.e.}, the time cost of adaptive attack is around $133$ times as big as that of defense.

For the defense method described in Sec. \ref{DPC}, since two diffusion paths should be maintained during purification, the time complexity is simply doubled than the one in a single path (Sec. \ref{Conditional RDP}). 
It is concluded that although the inference time of our proposed test-time defense methods is increased, the increased cost also complicates the adaptive attacks as well.

Table \ref{tab: time cost comparison} shows the time cost comparison between the reverse diffusion process (purification) and adaptive attack (BPDA+$20$ EOT), implemented using the adjoint strategy in Sec. \ref{Sec: GGA}, under DiffPure and our defense methods. Here, we do not consider the adaptive {\AA} since the mechanism is much more complex beyond the above analysis.
% , where the ratio of Sec \ref{Conditional RDP} and \ref{DPC} is around $149$ and $115$, respectively.

% \begin{SCtable}[1.0]
%  \resizebox{0.5\textwidth}{!}{

\begin{table}[ht]
  \centering
  \resizebox{0.6\textwidth}{!}{
 \begin{tabular}{@{}cccc@{}}
   \toprule
   Methods & Purification Time (sec.) & Attack Time (sec.) & Ratio\\
   \midrule
   % DiffPure \cite{nie2022diffusion} & 7.24±1.05 & 693.28±57.05 & 95.76\\
   DiffPure \cite{nie2022diffusion} & 35.20±0.67 & 2508.56±116.06 & 71.27\\
   Sec. \ref{Conditional RDP} & 34.59±0.95 & 2504.14±112.61 & 72.39\\
   Sec. \ref{DPC} & 72.84±34.41 & 7517.03±1970.35 & 103.20 \\
   \bottomrule
 \end{tabular}}
 \vspace{5pt}
 \captionof{table}{Computational time cost comparison in five runs. This evaluation utilizes four testing images with batch size two, running on one NVIDIA V100.
 The ratio is calculated as the average time required for BPDA+20 EOT divided by that for reverse diffusion.}
 \label{tab: time cost comparison}
\end{table} 

\section{More Experimental Results}\label{Sec: More Results}
In this section, we evaluate the robustness of our method against several $\ell_2$-norm and black-box attacks. Also, we examine the robustness of our method using CIFAR-100 \cite{krizhevsky2009learning} and ImageNet \cite{deng2009imagenet}. 

\subsection{Results on More Attacks}
We tested our methods against $\ell_2$-norm optimization-based attacks and black-box attacks, including C\&W attack \cite{carlini2016towards}, SPSA \cite{uesato2018adversarial}, and the targeted black-box attack in both settings (nonadaptive/adaptive). We utilized the Python package, called \href{https://github.com/Harry24k/adversarial-attacks-pytorch}{Torchattacks}, to implement C\&W and SPSA attacks. The results are shown in  Table. \ref{Table: Robustness Evaluation Non-adaptive l2 & black-box}.
For the targeted black-box attack, we found that it is a special case of DiffAttack \cite{kang2024diffattack} since it only utilizes the input image and purified image to optimize the attack direction. The results are shown in Table. \ref{Table: Robustness Evaluation adaptive l2 & black-box & targeted}.
It can be observed that under these attacks, the robust accuracy is well maintained.

\begin{table}[ht]
  \centering
  \resizebox{0.43\textwidth}{!}{
  \begin{tabular}{@{}ccc@{}}
    \toprule
    Attacks & Clean Accuracy (\%) & Robust Accuracy (\%)\\
    \midrule[2pt]
    C\&W & 90.62±1.70 & 90.56±1.51 \\
    SPSA & 90.46±1.91 & 90.65±1.83 \\
    \bottomrule
  \end{tabular}}
  \vspace{10pt}
  \caption{Non-adaptive attacks on Sec. \ref{Conditional RDP}.
%  , including Anti-Adv \cite{alfarra2022combating}, DISCO \cite{ho2022disco}, DiffPure \cite{nie2022diffusion}, and GDMP \cite{wang2022guided,wu2022guided}. 
 % All comparisons were conducted under the same conditions (the same attack and the same NN model).
 Classifier: WRN-28-10. Dataset: CIFAR-10.}
  \label{Table: Robustness Evaluation Non-adaptive l2 & black-box}
\end{table}

\begin{table}[ht]
  \centering
  \resizebox{0.5\textwidth}{!}{
  \begin{tabular}{@{}ccc@{}}
    \toprule
    Attacks & Clean Accuracy (\%) & Robust Accuracy (\%)\\
    \midrule[2pt]
    Targeted attack & 95.31 & 95.31 \\
    C\&W & 90.17±4.31 & 85.09±6.37 \\
    SPSA & 89.32±4.85 & 89.19±4.81 \\
    \bottomrule
  \end{tabular}}
  \vspace{10pt}
  \caption{Adaptive attacks on Sec. \ref{DPC}.
%  , including Anti-Adv \cite{alfarra2022combating}, DISCO \cite{ho2022disco}, DiffPure \cite{nie2022diffusion}, and GDMP \cite{wang2022guided,wu2022guided}. 
 % All comparisons were conducted under the same conditions (the same attack and the same NN model).
 Classifier: WRN-28-10. Dataset: CIFAR-10.}
  \label{Table: Robustness Evaluation adaptive l2 & black-box & targeted}
\end{table}

\subsection{Results on CIFAR-100}
We provide the robustness evaluation against adversarial attacks on CIFAR-100, as shown in Table \ref{Table: Robustness Evaluation Cifar100} ({\em cf.} Tables \ref{Table: Robustness Evaluation Non-adaptive} and  \ref{Table: Robustness Evaluation Adaptive} for CIFAR-10).
Similarly, the experiments in the first block of Table \ref{Table: Robustness Evaluation Cifar100} are under the setting of non-adaptive attack, in which the attacker only knows the information of the downstream classifier. 
We also excerpt the results of \cite{rebuffi2021data, wang2023better, cui2023decoupled} from \href{https://robustbench.github.io/}{RobustBench} \cite{Robustbench2021} for more comparisons.
The second block of Table \ref{Table: Robustness Evaluation Cifar100} shows the results obtained under the setting of adaptive attacks.
Note that the results for DiffPure \cite{nie2022diffusion} are from \cite{zhang2024enhancing}.
We can find that our methods are either better than the prior works under BPDA+EOT or comparable with DiffPure under {\AA} and PGD-$\ell_\infty$. %{\color{blue} Especially, under BPDA+EOT attack, Sec. \ref{DPC} has better performance than Sec. \ref{Conditional RDP}. We believe that this is because the number of targeted images $x_1^{tar},\ldots,x_C^{tar}$ described in Sec. \ref{DPC} gets large from $10$ to $100$, so the raise of randomness makes it harder to attack.}

%%%On the other hand, we show the evaluation result on CIFAR-100 in terms of granularity in realizing {\AA} (Rand) ({\em i.e.,} one call vs. multiple calls of adjoint method), as described in Sec. \ref{Adv AutoAttack-Realization} and Sec. \ref{Sec: GGA}, in Table \ref{tab: AutoAttack diff-CIFAR-100} ({\em cf.} Table \ref{tab: AutoAttack diff} for CIFAR-10).
%%%We find again that if adversarial examples were generated from DiffPure+WRN-28-10 with the original code of DiffPure, the robust accuracy for ours and DiffPure is almost the same. 
%Nevertheless, by using the adversarial examples generated from our implementation of {\AA} (Rand) with multiple calls, both the robust accuracy drops but our method is still more robust than DiffPure.
%The evaluation results on CIFAR-100 verify our argument in Sec. \ref{Adv AutoAttack-Realization} again.

\begin{table*}[t]
  \centering
  \resizebox{0.8\textwidth}{!}{
  \begin{tabular}{@{}cccc@{}}
    \toprule
    Defense Methods & Clean Acc (\%) & Robust Acc (\%) & Attacks\\
    \midrule[2pt]
    No defense & 81.66 & 0 & PGD-$\ell_\infty$\\
    % \hdashline
    % AWP \cite{wu2020adversarial}* &  & & {\AA} (Standard)\\ 
    % Anti-Adv \cite{alfarra2022combating}* + AWP \cite{wu2020adversarial} &  &  & {\AA} (Standard)\\ \hdashline
    % DISCO \cite{ho2022disco}*&  &  & PGD-$\ell_\infty$\\
    %\cite{rebuffi2021data, wang2023better, cui2023decoupled}
    Rebuffi {\em et al.} \cite{rebuffi2021data} & 62.41 & 32.06 & {\AA} (Standard)\\
    Wang {\em et al.} \cite{wang2023better} & 72.58 & 38.83 & {\AA} (Standard)\\  Cui {\em et al.} \cite{cui2023decoupled} & \textbf{73.85} & 39.18 & {\AA} (Standard)\\
    DiffPure \cite{nie2022diffusion} & \textbf{61.96±2.26} & 59.27±2.95 &PGD-$\ell_\infty$\\
    DiffPure \cite{nie2022diffusion} & 61.98±2.47 & \textbf{61.19±2.87} &{\AA} (Standard)\\
    % SOAP \cite{shi2021online}*&  &  &PGD-$\ell_\infty$\\
    % Hill {\em et al}. \cite{hill2020stochastic}*&  &  &PGD-$\ell_\infty$\\
    % ADP ($\s=0.1$) \cite{yoon2021adversarial}*&  &  &PGD-$\ell_\infty$\\
    Ours (Sec. \ref{Conditional RDP}) %with $\eta=$ 2.5e-3) 
    & 61.71±2.49 & \textbf{60.21±1.82} & PGD-$\ell_\infty$\\
    % Ours (Sec. \ref{DPC}) %with $\eta=$ 2.5e-3) 
    % & 59.25±2.57 & 58.42±2.70 & PGD-$\ell_\infty$\\ \hdashline
    Ours (Sec. \ref{Conditional RDP}) %with $\eta=$ 2.5e-3) 
    & 62.02±2.24 & 60.08±2.44 & {\AA} (Standard)\\
    % Ours (Sec. \ref{DPC}) %with $\eta=$ 2.5e-3) 
    % & 59.92±2.51 & 57.85±3.35 & {\AA} (Standard)\\
    \midrule[2pt]
    No defense & 81.66 & 0 & BPDA+EOT\\
    DiffPure \cite{nie2022diffusion,zhang2024enhancing} & 69.92 & 48.83 &BPDA+EOT\\
    Hill {\em et al.} \cite{hill2020stochastic}*&  51.66 & 26.10 &BPDA+EOT\\
    ADP ($\s=0.1$) \cite{yoon2021adversarial}*& 60.66 & 39.72 & BPDA+EOT\\
    % Ours (Sec. \ref{Conditional RDP}) %with $\eta=$ 2.5e-3) 
    % & 73.29±2.94 & 50.54±4.05 & BPDA+EOT\\
    Ours (Sec. \ref{DPC}) %with $\eta=$ 2.5e-3) 
    & \textbf{70.38±4.05} & \textbf{51.25±3.82} &BPDA+EOT\\
    \bottomrule
  \end{tabular}}
  \caption{Robustness evaluation and comparison between our method and state-of-the-art methods. Classifier: WRN-28-10. Testing dataset: CIFAR-100. Asterisk (*) indicates that the results were excerpted from the papers. Boldface indicates the best performance for each attack.}
  \label{Table: Robustness Evaluation Cifar100}
\end{table*}

%\begin{table}[h] 
%  \centering
%  \normalsize
%  \renewcommand\arraystretch{1.3}
%  \resizebox{0.5\textwidth}{!}{
%  \begin{tabular}{@{}ccc@{}}
%    \toprule
%    Defenses & Adv from \cite{nie2022diffusion} & Adv from Ours\\
%    \midrule
%    DiffPure \cite{nie2022diffusion}& 26.56\% & 20.31\%  \\
%    Ours & 26.56\%  & 25.00\%  \\
%    \bottomrule
%  \end{tabular}}
%  \caption{Robust accuracy for adversarial examples generated from different implementations of diffusion purification ({\em i.e.}, Ours and DiffPure). Our implementation uses output in every time step from \code{torchsde}, whereas DiffPure uses \code{torchsde} without accessing the intermediate outputs, which is encapsulated in \code{torchsde} function call. Adv denotes adversarial sample obtained from adaptive {\AA}  (Rand) with $20$ EOT. Classifier: WRN-28-10. Dataset: Subset of CIFAR-100.}
%  \label{tab: AutoAttack diff-CIFAR-100}
%\end{table}

\subsection{Results on ImageNet}
We provide the robustness evaluation against non-adaptive PGD-$\ell_\infty$ on ImageNet, as shown in Table \ref{tab: Robustness Evaluation ImageNet}. This experiment was conducted on a more advanced transformer-based classifier \cite{liu2022swin} with $\norm{\delta}_\infty\le8/255$ and ResNet-50 with $\norm{\delta}_\infty\le4/255$.
Our method proposed in Sec. \ref{DPC} is slightly better than DiffPure, while Our method proposed in Sec. \ref{Conditional RDP} is comparable with DISCO, which, however, needs additional data for training EDSR for purification.

\begin{table}[ht]
  \centering
  \resizebox{0.8\textwidth}{!}{
  \begin{tabular}{@{}ccccc@{}}
    \toprule
    Defense Methods & Clean Acc (\%) & Robust Acc (\%) & Classifier & Attack\\
    \midrule[2pt]
    % No defense &  & $0$ & PGD-$\ell_\infty$\\
    %DiffPure & 75.13±11.67 & 73.11±11.76 &PGD-$\ell_\infty$\\
%    DiffPure [31] & 64.39±12.89 & 55.93±12.89 & ResNet-18 & \multirow{5}{*}{PGD-$\ell_\infty$}\\
%    Sec. 4.2, $\eta=$ 2.5e-3 & 64.39±13.01 & 53.85±12.43 & ResNet-18 & \\
    DISCO \cite{ho2022disco}* &\textbf{72.64} & 66.32&ResNet-50&PGD-$\ell_\infty$ ($4/255$)\\
    Ours (Sec. \ref{Conditional RDP})%, $\eta=$ 2.5e-3 
    & 69.32±12.18 & \textbf{68.12±12.13} & ResNet-50& PGD-$\ell_\infty$ ($4/255$)\\
    Ours (Sec. \ref{DPC})%, $\eta=$ 2.5e-3 
    & 67.55±11.73 & 66.54±12.15 & ResNet-50& PGD-$\ell_\infty$ ($4/255$) \\
    DiffPure \cite{nie2022diffusion} & 75.13±11.67 & 73.11±11.76 & SwinV2 \cite{liu2022swin}  & PGD-$\ell_\infty$ ($8/255$)\\
    Ours (Sec. \ref{Conditional RDP})%, $\eta=$ 2.5e-3 
    & 75.37±12.01 & 71.73±12.63 & SwinV2 \cite{liu2022swin} & PGD-$\ell_\infty$ ($8/255$)\\
    %Sec. 4.3, $\eta=$ 2.5e-3 & 73.17±11.20 & 72.29±11.92 & SwinV2 & \\
    Ours (Sec. \ref{DPC})%, $\eta=$ 1.0e-3 
    & \textbf{75.38±12.78} & \textbf{73.20±11.27} & SwinV2 \cite{liu2022swin} & PGD-$\ell_\infty$ ($8/255$)\\
    % Ours (Sec. \ref{Conditional RDP} with $\eta=$ 2.5e-3) & 70.33±11.82 & 66.16±14.03 & {\AA} (Standard)\\
    % Ours (Sec. \ref{DPC} with $\eta=$ 2.5e-3) & 66.48±12.19 & 63.32±13.32 & {\AA} (Standard)\\
    \bottomrule
  \end{tabular}}
  %\vspace{-3.5mm}
  \vspace{10pt}
  \caption{Robustness comparison between our method and DISCO/DiffPure. Testing dataset: ImageNet. Asterisk (*) indicates that the results were excerpted from the paper. Boldface indicates the best performance for each attack.}
  %\label{Table: Robustness Evaluation ImageNet}
  % \vspace{-8.5mm}
  \label{tab: Robustness Evaluation ImageNet}
\end{table}

\end{document}